\documentclass{ieeeaccess}

\usepackage{amsmath,amssymb,amsfonts}
\usepackage{xcolor}
\usepackage{diagbox}
\usepackage{algorithmic}
\usepackage{graphicx}
\usepackage{textcomp}
\usepackage{array,multirow}
\usepackage{float}

\def\bg#1{\mbox{\boldmath$#1$}}

\usepackage{url}
\usepackage{arydshln}
\usepackage{cite}
\usepackage{xcolor}

\begin{document}
\history{Received 20 September 2022, accepted 6 October 2022, date of publication 10 October 2022, date of current version 18 October 2022.}
\doi{10.1109/ACCESS.2022.3213652}

\title{ViGAT: Bottom-up event recognition and explanation in video using factorized graph attention network}

\author{\uppercase{Nikolaos~Gkalelis},
\uppercase{Dimitrios~Daskalakis, and Vasileios~Mezaris},
\IEEEmembership{Senior~Member,~IEEE}}
\address{Information Technologies Institute/Centre for Research and Technology Hellas (CERTH), Thermi 57001, Greece (email: gkalelis@iti.gr; dimidask@iti.gr; bmezaris@iti.gr).}
\tfootnote{This work was supported by the EU Horizon 2020 programme under grant agreements 832921 (MIRROR) and 101021866 (CRiTERIA); and, by the QuaLiSID - ``Quality of Life Support System for People with Intellectual Disability'' project, which is co-financed by the European Union and Greek national funds through the Operational Program Competitiveness, Entrepreneurship and Innovation, under the call RESEARCH-CREATE-INNOVATE (project code: T2EDK-00306).}

\markboth
{N. Gkalelis \headeretal: ViGAT: Bottom-up event recognition and explanation in video using factorized graph attention network}
{N. Gkalelis \headeretal: ViGAT: Bottom-up event recognition and explanation in video using factorized graph attention network}

\corresp{Corresponding author: Vasileios Mezaris (e-mail: bmezaris@iti.gr).}

\begin{abstract}
In this paper a pure-attention bottom-up approach, called ViGAT, that utilizes an object detector together with a Vision Transformer (ViT) backbone network to derive object and frame features, and a head network to process these features for the task of event recognition and explanation in video, is proposed.
The ViGAT head consists of graph attention network (GAT) blocks
factorized along the spatial and temporal dimensions in order to capture effectively both local and long-term dependencies between objects or frames.
Moreover, using the weighted in-degrees (WiDs) derived from the adjacency matrices at the various GAT blocks, we show that the proposed architecture can identify the most salient objects and frames that explain the decision of the network.
A comprehensive evaluation study is performed, demonstrating that the proposed approach provides state-of-the-art results on three large, publicly available video datasets (FCVID, MiniKinetics, ActivityNet)\footnote{Source code is made publicly available at: \url{https://github.com/bmezaris/ViGAT}
}.
\end{abstract}

\begin{keywords}
Video event recognition, eXplainable AI (XAI), graph attention network, factorized attention, bottom-up.
\end{keywords}

\titlepgskip=-15pt

\maketitle

\section{Introduction}
\label{sec:intro}

\begin{figure}[!htb]
\begin{center}
\begin{tabular}{cc}
\includegraphics[width=.45\columnwidth]{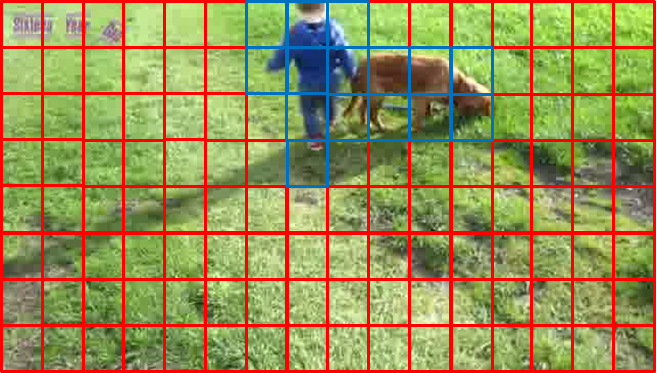} &
\includegraphics[width=.45\columnwidth]{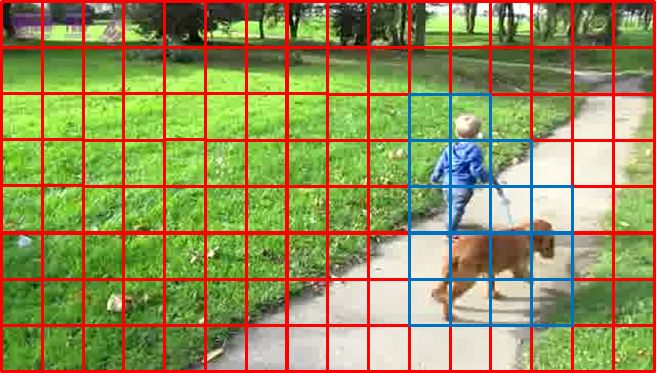} \\
\includegraphics[width=.45\columnwidth]{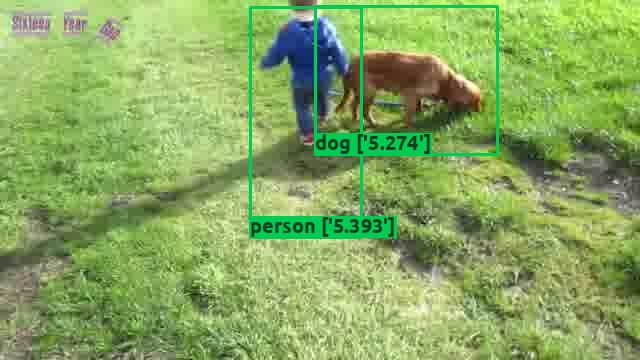} &
\includegraphics[width=.45\columnwidth]{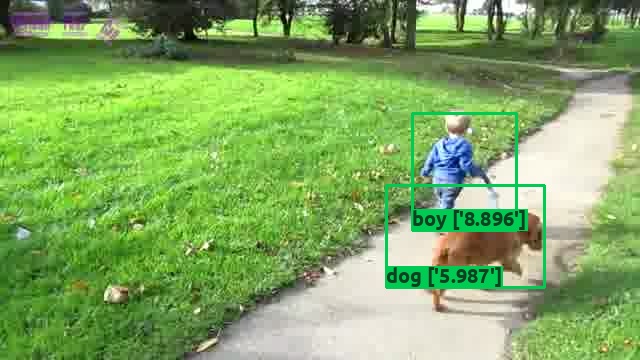}
\end{tabular}
\end{center}
\caption{Illustration of how top-down (1st row) and bottom-up (2nd row) approaches learn to focus on the salient frame regions, using a video labelled as ``Walking the dog'' event.
Top-down approaches explicitly (e.g. Transformers) or implicitly (e.g. CNNs) ``patchify'' each frame to generate patch proposals in a context-agnostic way;
the video labels are then used to train the network so that it learns to focus on the patches mostly related with the event (e.g. the 32 blue patches in this example) while ignoring the rest of them (the red patches).
Instead, the proposed bottom-up approach supports the classifier by providing the main objects depicted in the frames.
Such an approach can also facilitate the generation of object- and frame-based explanations about the event recognition outcome.
An example of this is shown in the second row of the figure.}
\label{fig:Motiv}
\end{figure}

\PARstart{D}{ue} to the explosion in the creation and use of video data in many sectors, such as entertainment and social media, to name a few, there is a great demand for analyzing and understanding video content automatically.
Towards this direction, the recognition of high-level events and actions in unconstrained videos plays a crucial role for improving the quality of provided services in various applications, e.g. \cite{JiangIEEEAccess2022,VieiraIEEEAccess2022,MemonIEEEAccess2021,WangCASVT2021,GkalelisCVPR2021,GhodratiCVPR2021,Meng20}.

The introduction of deep learning approaches has offered major performance leaps in video event recognition \cite{QiTCSVT2020,GkalelisCVPR2021,GhodratiCVPR2021,Meng20,WangECCV18,Zhang_2021_ICCV,bertasius_2021_ICML,Arnab_2021_ICCV,fan2021multiscale,Girdhar_2021_ICCV}.
Most of these methods operate in a top-down fashion \cite{GhodratiCVPR2021,Meng20,Zhang_2021_ICCV,bertasius_2021_ICML,Arnab_2021_ICCV,fan2021multiscale,Girdhar_2021_ICCV}, i.e. they utilize a network architecture to directly extract patch-, frame- or snippet-level features; and, through an appropriate loss function (e.g cross-entropy), exploit the class labels to learn implicitly the video regions that are mostly related with the specified action or event.
For instance, state-of-the-art Transformers \cite{Zhang_2021_ICCV,Arnab_2021_ICCV,Girdhar_2021_ICCV} segment image frames using a uniform grid to produce a sequence of patches, as shown in the first row of Fig. \ref{fig:Motiv}.
A similar image partitioning is also imposed implicitly by convolutional neural networks (CNNs), where the patch size is determined by the CNN's receptive field \cite{Zhang_2021_ICCV}.
This ``patchifying'' is context-agnostic and usually only a small fraction of the patches contains useful information about the underlying event.
During the supervised learning procedure the Transformer or CNN learns to disregard patches irrelevant to the target event, while extracting and synthesizing information from the patches that are related to the target event.
Considering that the real action or event may be occurring in only a small spatiotemporal region of the video, this procedure is expensive; it is also suboptimal to start by treating all image patches equally, as a large amount of them is irrelevant and does not need to be thoroughly analyzed \cite{Sydorov_2019_BMVC,Chi_CVPR_2020,LiCVPR21,Li_2022_WACV,WangCVPR22}.

Studies in cognitive science suggest that humans interpret complex scenes by selecting a subset of the available sensory information in a bottom-up manner, most probably in order to reduce the complexity of scene analysis \cite{TsotsosArtifIntel1995,IttiPAMI1998,Chi_CVPR_2020}.
It has also been shown that the same brain area is activated for processing object and action information for recognizing actions \cite{NelissenScience2005,GuptaPAMI2009}.
Finally, psychological studies suggest that events may be organized around object/action units encoding their relations, and that this structural information plays a significant role in the perception of events by humans \cite{ZacksJEPG2001,KurbyTCS2008,JiCVPR2020}.

Motivated by cognitive and psychological studies as described above, recent bottom-up action and event recognition approaches \cite{GkalelisCVPR2021,WangECCV18} represent a video frame using not only features extracted from the entire frame but also features representing the main objects of the frame.
More specifically, they utilize an object detector to derive a set of objects depicting semantically coherent regions of the video frames, a backbone network to derive a feature representation of these objects, and an attention mechanism combined with a graph neural network (GNN) to classify the video.
In this way, the classifier is supported to process in much finer detail the main video regions that are expected to contain important information about the underlying event \cite{Sydorov_2019_BMVC}.
The experimental evaluation in these works has shown that the bottom-up features constitute strong indicators of the underlying events and are complementary to the features extracted from the entire frames.
More specifically, in \cite{WangECCV18}, an I3D video backbone model is applied to extract spatiotemporal features, object proposals are generated using RoIAlign \cite{He_2017_ICCV}, an attention mechanism \cite{Vaswani_2017_NIPS} is used to construct the adjacency matrix of the spatiotemporal graph whose nodes are the object proposals, and a GNN is used to perform reasoning on the graph.
However, the use of 3D convolutions in the above work to represent the video may not be adequate for describing actions or events that require long-term temporal reasoning, as for instance is explained in {\cite{WangCVPR2018,Kaiyu_NIPS_2018,Chi_CVPR_2020,Zhang_2021_ICCV,bertasius_2021_ICML,Arnab_2021_ICCV,Girdhar_2021_ICCV}}.
Moreover, a large graph is constructed that captures the spatiotemporal evolution of the objects along the overall video, which imposes strict limitations in terms of memory requirements and also makes it difficult to sample a larger number of frames to improve recognition performance (see \cite{Arnab_2021_ICCV}: Fig. 7 and the related ablation study concerning the effect of the number of frames in the action recognition performance).
In \cite{GkalelisCVPR2021}, the 3D-CNN backbone of \cite{WangECCV18} is replaced by a 2D-CNN (i.e. ResNet \cite{HeCVPR2016}), and an attention mechanism \cite{YangCVPR2020} with a GNN are used to encode the bottom-up spatial information at each frame only; the sequence of feature vectors is then processed by an LSTM \cite{HochreiterNeuralComp97} to classify the video.
Therefore, in contrast to \cite{WangECCV18}, the above architecture factorizes the processing of the video along the spatial and temporal dimension, thus, effectively removing the memory restrictions imposed in \cite{WangECCV18} by the use of expensive 3D-CNN and the construction of the large spatiotemporal attention matrix.
Moreover, the authors in \cite{GkalelisCVPR2021} make a first attempt at exploiting the weighted in-degrees (WiDs) of the graph convolutional network's (GCN's) adjacency matrix to propose eXplainable AI (XAI) criteria and provide object-level (i.e., spatial) explanations concerning the recognized event \cite{GkalelisCVPR2021}.
However, despite the fact that this architecture can process long sequences of video frames, it is well known that the LSTM struggles to model long-term temporal dependencies \cite{WangCVPR2018,Kaiyu_NIPS_2018,Chi_CVPR_2020,Zhang_2021_ICCV,bertasius_2021_ICML,Arnab_2021_ICCV,Girdhar_2021_ICCV}.
Additionally, only qualitative results of ObjectGraphs' explanation approach are presented in \cite{GkalelisCVPR2021}.

Recently, pure-attention top-down approaches, i.e. methods that aggregate spatiotemporal information via stacking attention for modelling more effectively the long-term dependencies in videos, have achieved superior video action recognition \cite{WangCVPR2018,Kaiyu_NIPS_2018,Chi_CVPR_2020,Zhang_2021_ICCV,bertasius_2021_ICML,Arnab_2021_ICCV} or activity anticipation \cite{Girdhar_2021_ICCV} performance over previous methods that use CNN or LSTM layers in their processing pipeline.
In this work, inspired by the above findings and building on the bottom-up approach of \cite{GkalelisCVPR2021}, we replace the hybrid GNN-LSTM head of \cite{GkalelisCVPR2021} with a graph attention network-based (GAT-based) head network to process both the spatial (object) features as well as the sequence of features derived from the multiple frames.
Our resulting head network, called hereafter ViGAT head, utilizes attention along both the spatial and temporal dimensions to process the features extracted from the video.
Moreover, we use the Vision Transformer (ViT) as backbone (instead of a ResNet backbone, used in \cite{GkalelisCVPR2021}) to derive a feature representation of both the frames and the detected objects. 
Therefore, in our work attention is factorized along three dimensions, i.e., i) spatially among patches within each object (by using ViT), ii) among objects within each frame, and iii) temporally along the video.
Thus, in overall, due to the use of the ViT backbone (instead of a ResNet one) and the employment of a fully-attention head (instead of a hybrid attention-LSTM one), the proposed ViGAT can extract much richer features and model more effectively the long-term dependencies of video events in comparison to ObjectGraphs \cite{GkalelisCVPR2021}.
Additionally, in contrast to \cite{GkalelisCVPR2021}, which learns an adjacency matrix with respect to the objects at individual frames, and can thus derive only object-based explanations, we also derive an adjacency matrix along the temporal dimension, i.e. with respect to individual frames.
Thus, the WiDs calculated from the different learned adjacency matrices in the ViGAT head  (i.e. along the spatial and temporal dimensions) facilitate the derivation of multilevel explanations regarding the event recognition result, i.e., the extraction of not only the salient objects but also of the most salient frames explaining the model's outcome.
We should also note that despite the fact that the extraction of bottom-up (object) information increases the computational complexity of the proposed approach, during training this is only done once using the pretrained object detector and ViT backbone; thus, compared to the majority of other methods, which typically train the employed backbone end-to-end along with the rest of their components, ViGAT has a significantly lower training complexity.
Finally, following other works in the literature \cite{InanICLR2017,Lan_2020_ICLR,GuohaoICML2021}, we also explore the weight-tying of the individual GAT blocks in the ViGAT head of the proposed model to further reduce its memory footprint.
Extensive experiments demonstrate that the proposed approach provides state-of-the-art performance on three popular datasets, namely, FCVID \cite{FCVID}, MiniKinetics \cite{XieECCV18} and ActivityNet \cite{caba2015activitynet}.
Summarizing, our main contributions are the following:
\begin{itemize}
\item We propose the first, to the best of our knowledge, bottom-up pure-attention approach for video event recognition. A ViT backbone derives feature representations of the objects and frames, obtaining rich bottom-up information about the video scenes; and, an attention-based network head (called ViGAT head) is factorized along the spatial and temporal dimensions in order to identify the most interesting scene parts and thus capture effectively the long-term dependencies of events in video.
\item  We contribute to the field of explainable AI by demonstrating how to exploit the WiDs of the adjacency matrices at the various levels of the ViGAT head in order to derive explanations along the spatial and temporal dimensions for the event recognition outcome; and, by successfully adapting popular XAI measures from the image recognition domain, being the first to quantitatively document the goodness of temporal (frame) explanations for video event recognition.
\end{itemize}
The structure of the paper is the following:
Section \ref{sec:RelatedWork} presents the related work.
The proposed method is described in Section \ref{sec:ProposedMethod}.
Experimental results are provided in Section \ref{sec:ExperimentalEvaluation} and
conclusions are drawn in Section \ref{sec:Conclusions}.

\section{Related work}
\label{sec:RelatedWork}

\subsection{Video event and action recognition}
\label{ssec:VideoRecognition}
In this section, a survey of deep-learning-based video event and action recognition approaches is presented.
For a broader literature survey on this topic the interested reader is referred to \cite{ZhuArxiv2020,PareekASO2021}.

\subsubsection{Top-down approaches}
\label{sssec:TopDown}
The majority of event and action recognition approaches are top-down. We further categorize these methods according to their design choices in relation to feature extraction.

\textit{Convolutional 2D.} These approaches utilize architectures with 2D convolutional kernels to extract features at frame-level.
In \cite{Simonyan14}, a two-stream network is proposed that utilizes a spatial and a temporal branch to process independently RGB and optical flow frames.
This architecture can utilize deep CNNs pretrained on large-scale datasets, but can only operate on single frames and is computationally expensive due to performing dense video sampling.
TSN \cite{WangECCV16} extends the above work extracting sparsely sampled snippets, i.e. dividing the video to a few segments of equal length and selecting randomly one frame from each segment, yielding a significantly lower computational cost.
The above techniques operate on frame-level to derive a classification score; then, simple late fusion, i.e. average pooling of these scores, is applied to classify the video.
Average pooling, however, ignores the temporal ordering  and other higher-order rich statistical information which is useful for capturing complex dynamics of actions in video.
To go beyond late fusion, in \cite{LiAAAI19}, a factorized bilinear operator is incorporated into the network's convolutional layers to capture pairwise interactions among CNN features of adjacent frames and utilize more effectively the temporal relations across frames.
In \cite{Kaiyu_NIPS_2018}, the non-local module \cite{WangCVPR2018} (which is a kind of self-attention mechanism for modeling the correlation between spatial positions in feature maps) is generalized to model the interactions between positions across channels in ResNet backbones, resulting in a modified backbone that captures more effectively the long-term dynamics of actions in videos.
In \cite{GaoNIPS21}, a new attentive polling mechanism is integrated in various CNN backbone networks to combine frame-level action recognition scores.
In \cite{ZhaoCASVT2018}, VLAD pooling \cite{Jegou_CVPR_2010} that has shown state-of-the-art performance in combining hand-crafted features, is utilized to aggregate the features derived from the temporal and spatial CNN-based streams.
In \cite{GirdharCVPR17}, ActionVLAD derives a global feature descriptor for the entire video, using learnable pooling (NetVLAD \cite{ArandjelovicGTP18}) aggregating both appearance and temporal information along the video.
In contrast to the above works, where the exact temporal ordering of the descriptors is ignored, spatiotemporal VLAD (ST-VLAD)  \cite{SoltanianAG20} reformulates the VLAD optimization problem using Lagrange multipliers to
impose the minimization of the difference between the VLAD descriptors corresponding to neighboring frames.
As a result, the derived VLAD descriptors of the video signal vary smoothly along the temporal dimension.
Similarly, in \cite{SongTCSVT2020}, a 2D descriptor, called VideoMap, which is a row-wise layout of the per-frame vectorized CNN's features, is learned, for action classification.
Several works have also used recurrent neural networks to process the extracted CNN time series features.
In \cite{XuICMEW18}, a pretrained ResNet is used to derive a feature representation for each frame and an LSTM to process the temporal information.
In \cite{MemonIEEEAccess2021}, a 2D-CNN and LSTM are used to process the spatiotemporal video information, and in addition, shot boundary detection is applied to segment and predict multiple actions occurring in a video.
PivotCorrNN \cite{SunghunECCV18} introduces contextual gated recurrent units (cGRUs) to exploit time-varying information among different modalities (MFCC, IDT, etc.).
Although many of the above approaches utilize rather sparsely sampled frames, the extraction of a feature representation for each sampled frame using a rather deep CNN is still a computationally expensive process.

In response to the above drawback, techniques that use reinforcement learning and/or a gating network in order to further reduce the number of video frames being processed have also emerged.
In \cite{WuCVPR19}, AdaFrame exploits a policy gradient method to select future frames for faster and more accurate video predictions.
In \cite{WuICCV19}, a frame sampling strategy is learned using multi-agent reinforcement learning (MARL).
In \cite{GaoCVPR2020}, instead of a complex reinforcement learning policy network,
ListenToLook introduces the audio-modality to build a video skimming mechanism for selecting the most salient clips for the recognition task.
The above approaches utilize a fixed size network (i.e. with fixed memory footprint) irrespectively of video's complexity.
In contrast, LiteEval \cite{WuNIPS19} determines dynamically the frame resolution and utilizes a coarse- and a fine-LSTM cooperating through a binary gating module that decides whether additional high-resolution frames are necessary, thus leveraging network capacity dynamically.
Furthermore, the adaptive resolution network (AR-Net) \cite{Meng20}, instead of an expensive reinforcement learning mechanism or an additional audio modality, utilizes a lightweight policy network that learns to compute the optimal frame resolution on-the-fly, allowing the recognition of multiple video actions efficiently.
Contrarily to the above, AdaFocus \cite{Wang_2021_ICCV} utilizes a reinforcement learning policy network to leverage spatial redundancy, i.e., selects the most salient regions in the video frames with respect to the action recognition task.
In \cite{WangCVPR22}, AdaFocusV2 extends \cite{Wang_2021_ICCV} by replacing reinforcement learning with a differentiable interpolation-based patch selection operation, enabling efficient end-to-end optimization.
The above methods operate on untrimmed videos (i.e., videos that contain many irrelevant frames to the underlying action), where it is much easier to identify and discard less-significant image regions or entire frames.
In \cite{GowdaAAAI21}, differently from the above methods, the so-called SMART approach leverages a multi-frame attention and relation network to select the most informative frames in short trimmed videos.
In another line of research in the efficient video recognition paradigm, in \cite{VieiraIEEEAccess2022}, a low-cost CNN implemented in an embedded platform is used for violence recognition in video.

Regardless of whether the emphasis is on exploiting temporal and other statistical information or on improving efficiency, none of the top-down methods discussed in this section extracts and uses representations at a finer-than-frame level (e.g. for individual objects within a frame).

\textit{Convolutional 3D.} This category includes approaches with 3D convolutional kernels in the network architectures, operating at clip- or entire-video level.
C3D+LSVM \cite{TranICCV15} is one of the first works demonstrating that 3D convolutional kernels constitute a good descriptor for action recognition in video.
In \cite{CarreiraCVPR17}, a two stream architecture called I3D, which combines an optical flow and a 3D-CNN stream, is introduced. Additionally,  \cite{CarreiraCVPR17} describes how to leverage discriminant information from 2D-CNNs trained on ImageNet; it also shows that, when pretrained on a large-scale dataset (e.g. Kinetics), I3D provides recognition performance that is competitive to 2D-CNN approaches.
Both optical flow and 3D-CNN have high computational cost, limiting the applicability of the above two stream architectures in real-world applications \cite{XieECCV18}.
In order to reduce the computational overhead of the optical flow computation, \cite{CrastoCVPR19} employs a distillation approach during training to ingest optical flow stream information to a student network that operates on RGB frames.
Most works described above utilize relatively shallow networks, restricting the capacity of the networks for adequately learning a large number of complex video actions.
In \cite{HaraCVPR18}, ResNet-like 3D-CNN architectures of various depths are examined, with the authors concluding that carefully designed 3D-CNNs of large depth can improve the recognition performance when trained on large-scale datasets.
However, even when trained on large-scale datasets such heavy architectures can still suffer from overfitting.
To mitigate this problem, in \cite{KimCVPR20}, a multiplicative regularization approach, called random mean scaling, perturbs the low-frequency components of feature maps, effectively alleviating overfitting in deep 3D-ResNet architectures.
Similarly, in \cite{Chi_CVPR_2020}, a bilinear attentional mechanism (i.e. a bilinear matrix multiplication operator with learnable weights)  is introduced between network layers, extending the idea of non-local operators for 2D-CNNs \cite{WangCVPR2018} to the 3D-CNNs paradigm;
via directly connecting all locations of input feature maps, it is shown that this mechanism can capture the long-range spatiotemporal dynamics of video actions.
In SlowFast \cite{FeichtenhoferICCV2020}, a low- and a high-frame rate pathway, consisting of different-depth 3D-ResNets, are used to capture the spatial frame information and rapidly changing motion, respectively.
In \cite{HaIEEEAccess2021}, a 3D-CNN is first used to produce a feature representation for each video segment, which are then processed using an attention network with fast and slow pathways.
In \cite{LeeIEEEAccess2021}, 3D-CNN architectures are build using a temporal one-shot aggregation module to capture multiple temporal receptive fields, and depth-wise spatiotemporal factorized components for modeling short- and long-term motion dynamics.
In \cite{ZhouTMM2022}, a local and global branch are utilized using asymmetric convolution and two paralleled 1D-like convolutional blocks, to extract semantic and temporal action information,  respectively;
moreover, a supervised and self-supervised loss are combined to ingest information from labelled and unlabelled videos, respectively.
Contrarily to the above methods that leverage multi-scale spatiotemporal information,
in \cite{ZengIEEEAccess2021} a dynamic equilibrium module is inserted into a 3D-CNN backbone
to directly suppress the influence of spatiotemporal variations of actions in video.
In another line of research, in \cite{VuIEEEAccess2021},
a self-knowledge distillation approach is used to boost the performance of baseline 3D-CNN models 
(3D ResNet-18 and -50) for the task of action recognition.

Further to the above, several works also investigated how to reduce the high computational cost of using 3D-CNNs.
In \cite{TranCVPR2018}, separable 3D-CNNs are introduced, factorizing the 3D convolutional filters to a 2D spatial and a 1D temporal convolutional component, allowing faster processing of video sequences.
In \cite{XieECCV18}, the above work is further extended adding a feature gating mechanism, which is a simple self-attention operation.
In \cite{FayyazCVPR21}, a differentiable similarity guided sampling module is introduced in the architecture of 3D-CNNs
that measures the similarity of temporal feature maps and adaptively adjusts the temporal resolution.
In \cite{JiangIEEEAccess2022}, an efficient architecture is proposed, consisting of a 2D-CNN and two lightweight  1D-CNN-based  branches to capture spatial information, short- and long-term motion dynamics, respectively, and a 3D-CNN feature enhancement module to obtain more fine-grained spatial and temporal cues.
This architecture is much more efficient from SlowFast, which uses two 3D-ResNets in its branches.
SCSampler \cite{KorbarICCV9} extracts C3D features and as in \cite{GaoCVPR2020} for 2D-CNNs, the audio-modality is exploited to build a lightweight saliency model that selects short temporal clips within a long video that represent well the latter.
In another direction, a multigrid approach is proposed in \cite{YuanCVPR2020}, to derive variable mini-batch shapes (i.e. number of videos, frames and spatial resolution) during training, accelerating the training procedure and improving the generalization performance of 3D-CNNs.
In \cite{FeichtenhoferCVPR2020}, similarly to EfficientNet \cite{Tan_ICML_2019}, the X3D family of networks progressively expands a base network along different network dimensions (spatiotemporal resolution, frame rate, etc.) to  derive powerful  and efficient models.
In \cite{LiCVPR21}, Ada3D, trains a two-head network to learn frame and convolutional layer activation policies conditioned on the input video clip, thus reducing the computational cost of 3D-CNN models.
In \cite{GhodratiCVPR2021}, FrameExit utilizes X3D \cite{FeichtenhoferCVPR2020} for feature representation and applies a conditional early exiting to further improve the efficiency of the backbone network,
i.e., stops processing video frames when a sufficiently confident decision is reached.

In general, despite efforts to reduce the computational cost of using 3D-CNNs, such approaches typically continue to be much more expensive in terms of computational complexity and power consumption in comparison to their 2D-CNN counterparts.

\textit{Transformers}.
Convolutional or recurrent-based operations can only process a local neighborhood of the video in space and time; in order to model long-range dependencies, deep CNN or RNN architectures are utilized that stack several layers implementing the above operations, effectively extending the receptive field of the overall network.
However, the repetition of such local operations is computationally inefficient and causes optimization difficulties \cite{WangCVPR2018,Chi_CVPR_2020,YangICCV21}.
In contrast, Transformers utilize global self-attention to obtain a larger receptive field, thus, capturing more effectively the long-term dependencies in action videos \cite{YangICCV21}.
In \cite{DosovitskiyICLR21}, inspired from the success of Transformers in natural language processing, the so-called vision transformer (ViT) was introduced  outperforming convolutional-based approaches in popular image recognition benchmarks.
Subsequently, several attention-based architectures were also introduced concurrently for modeling the spatiotemporal contextual information of actions in videos \cite{Zhang_2021_ICCV,bertasius_2021_ICML,Arnab_2021_ICCV,Girdhar_2021_ICCV,fan2021multiscale}.
TimeSformer \cite{bertasius_2021_ICML} applies temporal and spatial attention,
demonstrating that in comparison to 3D convolutional networks
the attention-based architecture is faster and can be applied to much longer video clips.
Similarly, Video ViT (ViViT) \cite{Arnab_2021_ICCV} factorizes attention to spatial and temporal dimensions to efficiently process long video sequences  and proposes effective training strategies for ViViT by ablating different tokenization and regularization methods.
The spatiotemporal separable-attention video Transformer (VidTr) \cite{Zhang_2021_ICCV}
performs spatial and temporal attention separately and utilizes a deviation-based topK pooling operator to focus on the most representative frames of the video sequence.
In \cite{fan2021multiscale}, similarly to SlowFast \cite{FeichtenhoferICCV2020} and X3D \cite{FeichtenhoferCVPR2020} in the 3D convolutional paradigm, multiscale ViT (MViT) introduces several channel resolution scale stages into transformer models.

A common characteristic of all the above transformer-based approaches is that they rely on a context-agnostic extraction of a multitude of image patches, using a uniform grid, in order to learn the video actions; they do not take advantage of the inherent object-based composition of a visual scene and of the varying importance of specific objects for recognizing an event.

\subsubsection{Bottom-up approaches}
\label{sssec:BottomUp}
The methods described so far are top-down, i.e. entire frames or context-agnostic image patches corresponding to equally-sized receptive fields are processed along with the action class label of the video, to train a neural network to learn attending on the input frames or patches thereof that are related to the underlying action class.
Contrarily, bottom-up approaches utilize a more human-like mechanism to select a subset of the visual stimuli corresponding to salient image regions \cite{TsotsosArtifIntel1995,IttiPAMI1998}.
These methods typically use an object detector to
provide bottom-up information for training an event classifier \cite{QiTCSVT2020,LuoCASVT2021,GkalelisCVPR2021,WangECCV18,JiCVPR2020,WuCVPR2019}.
For instance, in \cite{WuCVPR2019}, a person detector (Faster R-CNN), a long-term feature bank and a 3D-CNN applied on short video segments, are used to provide long-term supportive information for action recognition.
In \cite{NigamCASVT2021} scene- and object-class pseudo-labels are derived for each video using pretrained networks (ResNet-50) on place365 and MS-COCO datasets respectively;
a multi-scale deformable 3D convolutional network and actor-object-scene attention model are then used for action recognition and factorization of actions into an actor, co-occurring objects, and scene cues.
In \cite{JiCVPR2020}, the Action Genome dataset is introduced, containing videos manually annotated with events, objects and their relationships, i.e., rich bottom-up information is provided, contrarily to \cite{NigamCASVT2021} where the objects are annotated at video clip level.
This dataset is used to learn a spatiotemporal scene graph feature-bank for action recognition;
during inference the Faster R-CNN and RelDN \cite{ZhangCVPR2019} are used to extract objects and visual relationships for building the spatiotemporal graph.
Since video object annotation is a labor-intensive and time consuming process, in \cite{QiTCSVT2020}, in contrast to \cite{JiCVPR2020} where object annotations are provided in the training set,
a region proposal network (R-FCN \cite{DaiNips2016}) and KLT trackers \cite{MunkresJSIAM1957} are used to derive and track video objects,
and build a semantic graph for each frame; subsequently, a hierarchical RNN is used to process the graph information and recognize group actions in video.
Instead of bounding boxes, semantic segmentation masks are extracted in \cite{LuoCASVT2021} using RefineNet-152;
this bottom-up information is combined with optical flow features derived using FlowNet2 in a two-stream architecture for the task of short-term action recognition.
In \cite{WangECCV18}, features extracted using a 3D-ResNet backbone with an object detector
(RoIAllign \cite{He_2017_ICCV}), are used to train an attention-based GNN,
which in comparison to RNNs or dense classification heads used above can learn more effectively the long-term dependencies of video actions.
In \cite{GkalelisCVPR2021}, object features are extracted at frame-level using an object detector with 2D-ResNet; these features are then used by a network head, composed of an attention mechanism, a GNN and an LSTM, factorizing the spatial and temporal dimension. 
In comparison to \cite{WangECCV18}, the above work factorizes the spatial and temporal dimension, allowing the efficient processing of long video signals.
Moreover, weighted in-degrees (WiDs) derived from the graphs' adjacency matrix are utilized to identify the most salient objects in the video that can explain the event recognition result.
Despite the considerable performance gains obtained by \cite{WangECCV18,GkalelisCVPR2021}, the use of 3D-CNN \cite{WangECCV18} or LSTM \cite{GkalelisCVPR2021} may not be adequate to fully capture the long-term dynamics of actions or events in video, as explained in \cite{Zhang_2021_ICCV,bertasius_2021_ICML,Arnab_2021_ICCV,Girdhar_2021_ICCV}.

In this work, to benefit from bottom-up video information while mitigating the above limitations, we propose a pure-attention bottom-up model utilizing an attention head network factorized along the spatial and temporal dimensions.
Additionally, using the temporal GAT components of our model, we are able to derive not only explanations at spatial level (i.e. objects, as in \cite{GkalelisCVPR2021})
but also at temporal level (i.e. frames).
Furthermore, we explore the possibility of tying the weights of the various GAT blocks to further reduce the memory footprint of the model, similarly to works in other domains \cite{InanICLR2017,Lan_2020_ICLR,GuohaoICML2021}.

\subsection{GNN decision explanation}
\label{ssec:VideoExplanation}

There have been only limited works studying the explainability of GNNs.
In contrast to CNN-based approaches where explanations are usually provided at pixel-level \cite{SudhakarSPJJK21}, for graph data the focus is on the structural information, i.e., the identification of the salient nodes and/or edges contributing the most to the GNN classification decision \cite{YuanExplainability}.
In the following, we briefly survey techniques most relevant to ours, i.e., targeting graph classification tasks and providing node-level (rather than edge-level) explanations.
For a broader survey of various works on explainability the interested reader is referred to \cite{YuanExplainability}.
In \cite{YingNIPS2019}, for each test instance the so-called GNNExplainer maximizes the mutual information between the GNN’s prediction and a set of generated subgraph structures to learn a soft mask for selecting the nodes explaining the model's outcome.
However, the explanation masks in  \cite{YingNIPS2019} are optimized individually for each input graph and thus may lack a global view \cite{YuanExplainability,AgarwalPMLR22}.
In \cite{VuNIPS20}, a surrogate, probabilistic graphical model that can learn the non-linear relationships of the input graph, as captured by the underlying GNN, is proposed.
More specifically, the so-called PGM-Explainer consists of a random perturbation approach to generate a synthetic dataset of graph data and respective predictions, a filtering step to discard unimportant graph data, and a learning step that trains the probabilistic graphical model using a Bayesian information criterion (BIC) score objective to provide explanations for the derived predictions.
Both the above approaches learn to derive explanations by minimizing an objective function -- mutual information \cite{YingNIPS2019} or BIC score \cite{VuNIPS20} -- whose relevance to explainability is unclear, as discussed in \cite{FunkeArxiv2021}.
Contrarily, in \cite{FunkeArxiv2021}, a new explainability measure called RDT-Fidelity is introduced, satisfying the desired properties of good explanations; subsequently, a combinatorial procedure called ZORRO is proposed
that uses a greedy forward selection algorithm to select the subgraphs that directly maximize the RDT-Fidelity score.
The approaches discussed above have shown promising results, however,
they introduce a high computational cost to the overall procedure, due to introducing an additional training step \cite{YingNIPS2019,VuNIPS20}
or because a greedy evaluation of a large number of possible node combinations is necessary \cite{FunkeArxiv2021}.
To this end, \cite{PopeCVPR19} extends popular gradient-based CNN methods to the GCN setting.
These methods are efficient as only one forward pass of the network is required; however, they suffer from the well-known gradient issues \cite{SaurabhWACV20}.

In this paper, to counter the described drawbacks of both gradient-based and computationally-expensive learning- or perturbation-based methods, we propose deriving WiD scores from the adjacency matrices at the various levels of the proposed attention head network; these WiD scores exhibit more stable behavior and improved explanation quality, and obtaining them introduces very limited computational overhead that is comparable to \cite{PopeCVPR19}.

\section{Video GAT}
\label{sec:ProposedMethod}

\begin{figure*}[!ht]
\centering
\includegraphics[width=1.65\columnwidth]{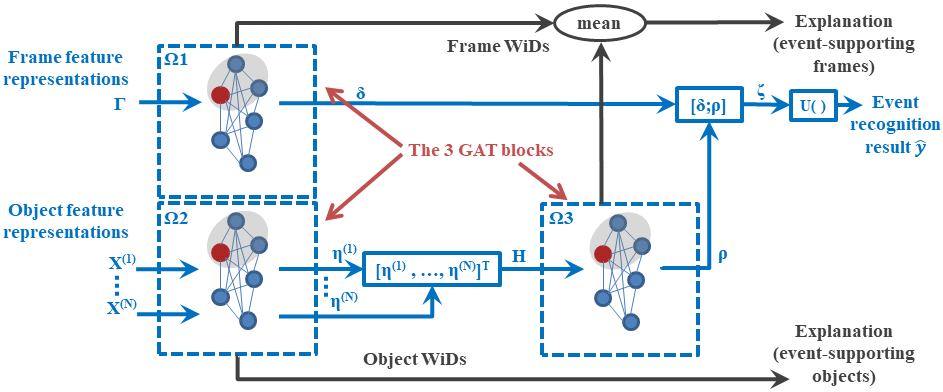}
\caption{Illustration of the proposed ViGAT head.
The GAT blocks $\Omega1$ and $\Omega2$ process the frame and object feature representations  \eqref{E:feaMatGlobal}, \eqref{E:feaMatLocal} at the input of the head.
The GAT block $\Omega3$ processes the new frame feature representations at the output of $\Omega2$.
The new video features at the output of the GAT blocks $\Omega2$ and $\Omega3$ are concatenated and the resulting feature is fed to layer $\mathsf{U}()$ to produce a score for each event class.
Additionally, the WiDs derived from the adjacency matrices of the three blocks provide comprehensive explanations (in terms of salient objects and frames) for the recognized event.}
\label{fig:Architecture}
\end{figure*}

\begin{figure}[!ht]
\centering
\includegraphics[width=0.99\columnwidth]{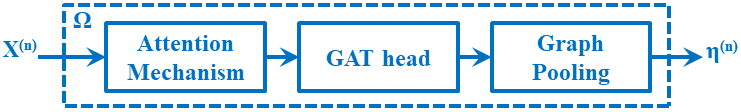}
\caption{GAT block and its components, i.e the attention mechanism \eqref{E:adjMatElm}, GAT head \eqref{E:DgcnLayer} and graph pooling \eqref{E:GraphPooling}.}
\label{fig:GATBlock}
\end{figure}

\begin{figure}[!ht]
\centering
\includegraphics[width=.737\columnwidth]{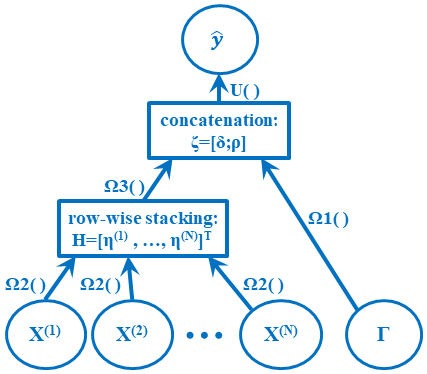}
\caption{Computational graph for learning the parameters of the ViGAT head (Fig. \ref{fig:Architecture}) using a dataset of videos represented as in \eqref{E:feaMatGlobal}, \eqref{E:feaMatLocal}.}
\label{fig:simpleExCmpGraph}
\end{figure}

\subsection{Video representation}
\label{ss:Problemformulation}

Let us assume an annotated video training set of $C$ event classes.
A video is represented with $N$ frames sampled from the video and a backbone network extracts a feature representation $\bg{\gamma}^{(n)} \in \mathbb{R}^F$ for each frame $n= 1, \dots, N$.
The feature representations are stacked row-wise to obtain matrix $\bg{\Gamma} \in \mathbb{R}^{N \times F}$,
\begin{equation}
\bg{\Gamma} = [\bg{\gamma}^{(1)}, \dots, \bg{\gamma}^{(N)}]^T. \label{E:feaMatGlobal}
\end{equation}
Similarly to recent bottom-up approaches \cite{WangECCV18,GkalelisCVPR2021}, we additionally use an object detector to derive $K$ objects from each frame; each object is represented by an object class label, a degree of confidence (indicating how confident the object detector is for this specific detection result), and a bounding box.
The backbone network is then applied to extract a feature representation $\mathbf{x}_k^{(n)} \in \mathbb{R}^F$ for each object $k$ in frame $n$. 
Sorting the feature representations in descending order according to their respective degree of confidence and stacking them row-wise we obtain the matrix $\mathbf{X}^{(n)} \in \mathbb{R}^{K \times F}$ representing frame $n$,
\begin{equation}
\mathbf{X}^{(n)} = [\mathbf{x}_1^{(n)}, \dots, \mathbf{x}_K^{(n)}]^T. \label{E:feaMatLocal}
\end{equation}
Although various backbones can be used, similarly to works in other domains, we use a Vision Transformer (ViT), which has shown excellent performance as backbone in a pure-attention framework \cite{Girdhar_2021_ICCV}.

\subsection{ViGAT head}
\label{ss:WeightTiedGcnBlocks}

The ViGAT head depicted in Fig. \ref{fig:Architecture} is used to process the features extracted from the backbone network.
It is composed of three GAT blocks, $\Omega1$, $\Omega2$ and $\Omega3$, where each block consists of a GAT and a graph pooling layer (the structure of the GAT block is described in detail in the next subsection).
Each GAT block is applied separately to a different feature type, effectively factorizing attention along the spatial and temporal dimensions.
This is a major advantage over the method of \cite{GkalelisCVPR2021}, where attention was utilized only along the spatial dimension;
the temporal video information was encoded using a less-effective LSTM structure.

More specifically, the feature representations of video frames \eqref{E:feaMatGlobal} and objects of frame $n$ \eqref{E:feaMatLocal} in the input of the GAT head are processed by the blocks $\Omega1$ and $\Omega2$, respectively, 
\begin{IEEEeqnarray}{rCl}
 \bg{\delta} &=& \mathsf{\Omega1}(\bg{\Gamma}), \\
 \bg{\eta}^{(n)} &=& \mathsf{\Omega2}(\mathbf{X}^{(n)}), \; n = 1, \dots, N, \label{E:blockOmega2}
\end{IEEEeqnarray}
where $\bg{\delta}, \bg{\eta}^{(n)} \in \mathbb{R}^{F}$ are new feature representations for the entire video and frame $n$, respectively.
Subsequently, the $N$ outputs of $\mathsf{\Omega2}$ (which correspond to the $N$ video frames) are stacked row-wise to obtain a new matrix $\mathbf{H} \in \mathbb{R}^{N \times F}$ for the overall video,
\begin{equation}
    \mathbf{H} = [\bg{\eta}^{(1)}, \dots, \bg{\eta}^{(N)}]^T. 
\end{equation}
This matrix is then fed to the block $\Omega3$ to obtain a second new feature representation $\bg{\varrho} \in \mathbb{R}^F$  for the entire video,
\begin{equation}
 \bg{\varrho} = \mathsf{\Omega3}(\mathbf{H}).
\end{equation}  
The derived features $\bg{\delta}$ and $\bg{\varrho}$ are then concatenated to form a new feature $\bg{\zeta} \in \mathbb{R}^{2F}$ for the video,
\begin{equation}
    \bg{\zeta} = [\bg{\delta}; \bg{\varrho}].
\end{equation}
Finally, $\bg{\zeta}$ is passed through a dense layer $\mathsf{U}()$ in order to derive a score vector $\mathbf{\hat{y}} = [\hat{y}_1, \dots, \hat{y}_C]^T$, where $\hat{y}_c$ is the classification score obtained for the $c$th event class.
Using an annotated training set, an appropriate loss function and learning algorithm, the ViGAT head can be trained end-to-end.
Moreover, in case that the weights of the three GAT blocks are tied (i.e. $\Omega1 = \Omega2 = \Omega3$), the gradient updates for the GAT block parameters are simply the sum of the updates obtained for the $N+2$ roles (see Fig. \ref{fig:simpleExCmpGraph}) of the GAT block in the network, as in \cite{InanICLR2017,GuohaoICML2021,ScarselliIEEETNN2009}.

\subsection{GAT block}
\label{ss:WeightTiedGcnBlocks}

The GAT block structure $\Omega$ depicted in Fig. \ref{fig:GATBlock} is the building block of the ViGAT head.
To avoid a notation clutter, we use in this section block $\Omega2$ \eqref{E:blockOmega2} as an example for defining the GAT block (blocks $\Omega1$, $\Omega2$, $\Omega3$ are identical).
The input to $\Omega2$ is matrix $\mathbf{X}^{(n)} \in \mathbb{R}^{K \times F}$ \eqref{E:feaMatLocal}, i.e. the feature representations of the $K$ objects of the $n$th frame.

The first component of the GAT block is an attention mechanism that is used to compute the respective matrix $\mathbf{E}^{(n)} \in \mathbb{R}^{K \times K}$ as follows \cite{Velickovic_ICLR_2018,YangCVPR2020,GkalelisCVPR2021},
\begin{IEEEeqnarray}{rCl}
\check{\mathbf{v}}_l^{(n)} &=& \check{\mathbf{W}} \mathbf{x}_l^{(n)} + \check{\mathbf{b}}, \label{E:adjMatAffineA} \\
\tilde{\mathbf{v}}_k^{(n)} &=& \tilde{\mathbf{W}} \mathbf{x}_k^{(n)} + \tilde{\mathbf{b}}, \label{E:adjMatAffineB} \\
e_{k,l}^{(n)} &=& \langle\check{\mathbf{v}}_k^{(n)}, \tilde{\mathbf{v}}_l^{(n)}\rangle, \label{E:adjMatSml} \end{IEEEeqnarray}
where, $\tilde{\mathbf{W}}$, $\check{\mathbf{W}} \in \mathbb{R}^{F \times F}$, $\tilde{\mathbf{b}}$, $\check{\mathbf{b}} \in \mathbb{R}^{F}$ are the weight matrices and biases of the attention mechanism, $\langle\,,\rangle$ is the inner product operator and $e_{i,j}^{(n)}$ is the attention coefficient at the $i$th row and $j$th column of $\mathbf{E}^{(n)}$.
The attention coefficients are then normalized across each row of $\mathbf{E}^{(n)}$ to derive the adjacency matrix $\mathbf{A}^{(n)} \in \mathbb{R}^{K \times K}$ of the graph \cite{Velickovic_ICLR_2018,WangECCV18,GkalelisCVPR2021,YangCVPR2020},
\begin{equation}
a_{k,l}^{(n)} = \frac{(e_{k,l}^{(n)})^2}{\sum_{\iota=1}^K (e_{k,\iota}^{(n)})^2}, \label{E:adjMatElm}
\end{equation}
where, $a_{k,l}^{(n)}$ is $\mathbf{A}^{(n)}$'s element at row $k$ and column $l$.

The derived adjacency matrix and the node features are then forwarded to a GAT head of $M$-layers \cite{WangECCV18,GkalelisCVPR2021,kipfIclr2017}
\begin{equation}
\mathbf{Z}^{[m]} = \sigma(\mathbf{A}^{(n)} \mathbf{Z}^{[m-1]} \mathbf{W}^{[m]}), \label{E:DgcnLayer}
\end{equation}
where, $m$ is the layer index (i.e. $m = 1, \dots, M$), $\sigma()$ denotes a nonlinear operation (here it is used to denote layer normalization \cite{ba2016layer} followed by element-wise $\mathsf{ReLU}$ operator), and
$\mathbf{W}^{[m]} \in \mathbb{R}^{F \times F}$, $\mathbf{Z}^{[m]} \in \mathbb{R}^{K \times F}$ are the weight matrix and output of the $m$th layer, respectively.
The input of the first layer is set to the input of the GAT block, i.e. $\mathbf{Z}^{[0]} = \mathbf{X}^{(n)}$, and the output of the GAT head, $\mathbf{\Xi}^{(n)} \in \mathbb{R}^{K \times F}$, is set to the output of its last layer, i.e. $\mathbf{\Xi}^{(n)} = \mathbf{Z}^{[M]}$.

Subsequently, graph pooling \cite{LeeIcml2019} is applied to produce a vector-representation of the graph at the output of the GAT block,
\begin{equation}
\bg{\eta}^{(n)} = \frac{1}{K} \sum_{k=1}^{K} \bg{\xi}_k^{(n)}, \label{E:GraphPooling}
\end{equation}
where $\bg{\xi}_k^{(n)} \in \mathbb{R}^{F}$ is the $k$th row of $\mathbf{\Xi}^{(n)}$.

We note that \eqref{E:DgcnLayer} resembles the layer-wise propagation rule of GCNs \cite{kipfIclr2017}.
However, as the exploitation of the attention mechanism to create the graph's adjacency matrix is central in our approach and due to the fact that this matrix is not symmetric (which violates the symmetry assumption in \cite{kipfIclr2017}), we resort to the more general message passing framework \cite{Justin_ICML_2017} and GAT \cite{Velickovic_ICLR_2018} to describe our model.

\subsection{ViGAT explanation}
\label{sec:MultiExplanations}

Considering that during the inference stage, the multiplication with the adjacency matrix in \eqref{E:DgcnLayer} amplifies the contribution of specific nodes, and the resulting video representation gives rise to the trained model's event recognition decision, the adjacency matrix can be used for deriving indicators of each node's importance in said model's decision.
This was first attempted in \cite{GkalelisCVPR2021}, where the importance of 
object $l$ at frame $n$ was estimated using the associated WiD value,
\begin{equation}
    \omega_2^{(l,n)} = \sum_{k=1}^{K} a_{k,l}^{(n)}, \label{E:ObjectWiD}
\end{equation}
where $a_{k,l}^{(n)}$ is $\mathbf{A}^{(n)}$'s element at $k$th row and $l$th column.

The qualitative results presented in \cite{GkalelisCVPR2021} demonstrated the usefulness of WiDs to produce explanations about the recognized video event.
However, the use of LSTM in \cite{GkalelisCVPR2021} to process the frame features restricted the computation of WiDs only to objects at static frames, and thus the derivation of explanations only at object-level.
In contrary, here we extend the utilization of WiDs in the temporal dimension.
Specifically, the use of temporal attention through blocks $\Omega1$ and $\Omega3$ to process the frame features enables us to derive two WiDs for the $n$th video frame,
\begin{IEEEeqnarray}{rCl}
    \omega_1^{(n)} &=& \sum_{\tau=1}^{N} \pi_{\tau,n}, \label{E:WiDGlobal} \\
    \omega_3^{(n)} &=& \sum_{\tau=1}^{N} \delta_{\tau,n}, \label{E:WiDLocal}
\end{IEEEeqnarray}
where, $\pi_{\tau,n}$, $\delta_{\tau,n}$ are the elements of matrices $\mathbf{\Pi} \in \mathbb{R}^{N \times N}$ and $\mathbf{\Delta} \in \mathbb{R}^{N \times N}$ at row $\tau$ and column $n$, and $\mathbf{\Pi}$, $\mathbf{\Delta}$ are the adjacency matrices of blocks $\Omega1$ and $\Omega3$, respectively (similarly to $\mathbf{A}^{(n)}$ being an adjacency matrix of block $\Omega2$, as computed in \eqref{E:adjMatElm}).
A large $\omega_1^{(n)}$ and/or $\omega_3^{(n)}$ indicates that the contribution of frame $n$ in the event recognition outcome is high.
In order to derive a single indicator for each frame, we average the above values to obtain a new indicator $\beta^{(n)}$ for the importance of frame $n$,
\begin{equation}
    \beta^{(n)} = \frac{1}{2}( \omega_1^{(n)} + \omega_3^{(n)} ). \label{E:widFrameMean}
\end{equation}
Equation \eqref{E:widFrameMean} is our proposed XAI criterion, i.e. we propose that the top-$\Upsilon$ frames with the highest $\beta^{(n)}$ values constitute an explanation of the network's event recognition outcome.

\section{Experiments}
\label{sec:ExperimentalEvaluation}

\subsection{Datasets}
\label{ssec:Datasets}

We run experiments on three large, publicly available event/action video datasets:
i) FCVID \cite{FCVID} is a multilabel video dataset consisting of 91223 YouTube videos annotated according to 239 categories.
It covers a wide range of topics, with the majority of them being real-world events.
The dataset is evenly split into training and testing partitions with 45611 and 45612 videos, respectively.
Among them, 436 videos in the training partition and 424 videos in the testing partition were corrupt and thus could not be used.
ii) MiniKinetics, which comes in two variants, one comprising approximately 130K video clips (121215 for training and 9867 for testing) \cite{Meng20} and one with approximately 85K clips (a 80K/5K training/testing split) \cite{XieECCV18}.
Both variants contain instances of 200 event/action classes and originate from the Kinetics dataset  \cite{KayArxiv2017}.
Each clip has been sampled from a different YouTube video, has 10 seconds duration and is annotated  with a single class label.
iii) ActivityNet v1.3 \cite{caba2015activitynet} is a popular multilabel video benchmark consisting of 200 classes (including a large number of high-level events), and 10024, 4926 and 5044 videos for training, validation and testing, respectively.
As the testing-set labels are not publicly available, the evaluation is performed on the so called validation set, as typically done in the literature.

\subsection{Setup}
\label{ssec:Setup}

Uniform frame sampling is one of the most commonly-used strategies in video action recognition due to its simplicity, efficiency and effectiveness, and has offered state-of-the-art results in this domain (e.g. see \cite{Simonyan14,WangECCV16,WangCVPR2018,GirdharCVPR17,SoltanianAG20,WuCVPR19,WuICCV19,Meng20,GhodratiCVPR2021,GowdaAAAI21} and references therein).
For this reason, uniform sampling is also applied here to represent each video with a sequence of $N$ frames in the input of the proposed ViGAT.

The number of sampled frames $N$ per video is selected based on the videos' average duration and the complexity of the actions in the respective dataset, also considering the number typically used in the relevant literature works.
The average duration of the videos in MiniKinetics and FCVID is 10 and 167 seconds, respectively   \cite{KayArxiv2017,FCVID}.
On the other hand, most videos in ActivityNet are much larger, i.e., with duration between 5 and 10 minutes \cite{caba2015activitynet}.
Concerning the complexity of the events/actions in the different datasets, FCVID mostly contains generic categories, such as ``baseball'', ``fire fighting'' and ``birthday''.
On the other hand, MiniKinetics and ActivityNet contain a broader variety spanning from high-level events to short-term actions that are more difficult to differentiate, such as ``applauding'' and ``clapping'', ``cleaning shoes'' and ``shining shoes'' (MiniKinetics), ``drinking beer'' and ``drinking coffee'', and ``long jump'' and ``triple jump'' (ActivityNet).
Based on the above analysis and following other works in the literature, we set $N$ to 9 frames for FCVID (e.g. as in \cite{WuCVPR19,Meng20,GowdaAAAI21,GkalelisCVPR2021}) and 30 frames for MiniKinetics (e.g. similarly to \cite{KimCVPR20,ZhouTMM2022}).
For ActivityNet, due to both video length and events complexity, we decided to sample a larger number of frames, i.e. $N=120$ (e.g. similarly to \cite{WuICCV19});
in this way, we want to ensure that the complex events/actions, especially the ones that resemble each other, as well as those covering only a small portion of the longer videos in this dataset, are adequately represented.

The object detector is used to extract a set of $K=50$ objects from each frame (the ones with the highest degree of confidence).
Thus, each object is represented with a bounding box, an object class label (which we only use for visualizing the object-level explanations) and an associated degree of confidence.
As object detector we use the Faster R-CNN \cite{renNips2015faster} with ResNet-101 \cite{HeCVPR2016} backbone, where feature maps of size $14 \times 14$ are extracted from the region of interest pooling layer.
The Faster R-CNN is pretrained and fine-tuned on ImageNet1K \cite{ILSVRC15} and Visual Genome \cite{krishna2017}, respectively.

ViGAT utilizes a pre-trained backbone network to derive a feature representation for each object in a frame as well as for the overall frame, as described in  \eqref{E:feaMatGlobal}, \eqref{E:feaMatLocal}.
We experimented with two backbones:
i) ViT: the ViT-B/16 variant of Vision Transformer \cite{DosovitskiyICLR21} pretrained on Imagenet11K and fine-tuned on Imagenet1K \cite{ILSVRC15} is our main backbone; specifically, the pool layer prior to the classification head output of the transformer encoder is used to derive a feature vector of $F=768$ elements,
ii) ResNet: a ResNet backbone is also used in order to compare directly with other literature works that use a ResNet backbone, and to quantify the performance improvement of the proposed pure-attention model (i.e. the effect of using attention also at object pixel-level through the ViT backbone); specifically, the pool5 layer of a pretrained ResNet-152 on ImageNet11K is used to derive an $F = 2048$ dimensional feature vector.

Concerning the ViGAT head (Fig. \ref{fig:Architecture}), the parameters of the three GAT blocks are tied, and $M=2$ layers \eqref{E:DgcnLayer} are used in each GAT head.
Moreover, $\mathsf{U}()$ is composed of two fully connected layers and a dropout layer between them with drop rate 0.5.
The number of units in the first and second fully connected layer is $F$ and $C$, respectively, where $C$ (the number of event classes) is equal to 239, 200 and 200 units, for the FCVID, MiniKinetics and ActivityNet dataset; 
the second fully connected layer is equipped with a sigmoid or softmax nonlinearity for the multilabel (FCVID, ActivityNet) or single-label (MiniKinetics) dataset, respectively.

We performed in total eight main experiments, one for each possible combination of dataset (FCVID, the two variants of MiniKinetics, ActivityNet) and backbone (ViT, ResNet).
In all experiments, the proposed ViGAT is trained using Adam optimizer with cross-entropy loss and  initial learning rate $10^{-4}$ (e.g. as in \cite{DosovitskiyICLR21}).
Following other works in the literature (e.g. \cite{Arnab_2021_ICCV}), a batch size of 64 is utilized, except for the experiment on ActivityNet with the ResNet backbone, where we reduced the batch size to 36 due to GPU memory limitations.
For the proposed ViGAT with ViT backbone the initial learning rate is multiplied by 0.1 at epochs 50, 90, for FCVID; 20, 50, for MiniKinetics; and 110, 160, for ActivityNet.
The total number of epochs is set to 100 for  MiniKinetics and 200 for FCVID and ActivityNet. 
For the ViGAT variant with ResNet backbone the initial learning rate is similarly reduced at epochs 30, 60; and 90 epochs are used in total for each dataset.
We should note that in all experiments
the proposed method exhibited a very stable performance with respect to different learning rate schedules. All experiments were run on PCs with an Intel i5 CPU and a single NVIDIA GPU (either RTX3090 or RTX2080Ti).

\subsection{Evaluation measures}
\label{ssec:Measures}

Similarly to other works in the literature and in order to allow for comparison of the proposed ViGAT with them, the event recognition performance is measured using the top-1 accuracy and mean average precision (mAP) \cite{ZhangEDS2009} for the single-label (MiniKinetics) and multilabel (FCVID, ActivityNet) datasets, respectively.

The explainability performance of ViGAT is measured using the top $\Upsilon$ frames of the video selected by it to serve as an explanation. 
We use two XAI evaluation measures used extensively for the explanation of CNN models, i.e., Increase in Confidence ($IC$) and Average Drop ($AD$) \cite{ChattopadhayWACV2018},
\begin{IEEEeqnarray}{rCl}
IC &=& \frac{1}{Q} \sum_{q=1}^{Q} \delta(\bar{y}_{q,\hat{u}_{q}} > \hat{y}_{q,\hat{u}_{q}}), \label{E:increaseinconf} \\
AD &=& \frac{1}{Q} \sum_{q=1}^{Q} \frac{max(0, \hat{y}_{q,\hat{u}_{q}} - \bar{y}_{i,\hat{u}_{q}})}{\hat{y}_{q,\hat{u}_{q}}}, \label{E:averagedrop}
\end{IEEEeqnarray}
where, $Q$ is the total number of evaluation-set videos, $\delta(a)$ is one when the condition $a$ is true and zero otherwise, $\hat{u}_{q}  \in \{1, \dots, C\}$ is the event class label estimated by the ViGAT model using all $N$ frames, $\hat{y}_{q,\hat{u}_{q}}$, $\bar{y}_{q,\hat{u}_{q}}$ are the model's scores for the $q$th video and estimated class $\hat{u}_{q}$, obtained using all or just the top $\Upsilon$ frames identified as explanations by the employed XAI criterion  \eqref{E:widFrameMean}, respectively.
That is, $IC$ is the portion of videos for which the model's confidence score increased, and $AD$ is the average model's confidence score drop, when just the $\Upsilon$ most salient frames are used to represent the video.
Higher $IC$ and lower $AD$ indicate a better explanation.
Additionally, we utilize two more general explainability measures, fidelity minus ($F-$) and fidelity plus ($F+$) \cite{YuanExplainability}, defined as
\begin{IEEEeqnarray}{rCl}
F- &=& \frac{1}{Q} \sum_{q=1}^Q (\delta(\hat{u}_{q} == u_{q}) - \delta(\bar{u}_{q} == u_{q})), \label{E:fidelityminus} \\
F+ &=& \frac{1}{Q} \sum_{q=1}^Q (\delta(\hat{u}_{q} == u_{q}) - \delta(\breve{u}_{q} == u_{q})), \label{E:fidelityplus}
\end{IEEEeqnarray}
where, $u_{q}$ is the ground truth label of the $q$th video, and $\bar{u}_{q}, \breve{u}_{q}$ are the labels estimated by the model using the top $\Upsilon$ (i.e., most salient) frames identified by our XAI criterion or the rest (i.e. the least salient) $N - \Upsilon$ frames,  respectively.
We see that $F-$ and $F+$ measure the impact on the model's performance when only the $\Upsilon$ most salient frames from each video are considered or are ignored, respectively.
Lower $F-$ and higher $F+$ denote a better explanation.

\subsection{Event recognition results}
\label{ssec:Results}

\begin{table}[!t]
\caption{Performance comparison on FCVID.}
\begin{center}
{
\begin{tabular}{lc}
       & mAP(\%) \\
\hline
  \text{ST-VLAD \cite{SoltanianAG20}}                 & 77.5 \\
  \text{PivotCorrNN \cite{SunghunECCV18}}             & 77.6 \\
  \text{LiteEval \cite{WuNIPS19}}                     & 80.0 \\
  \text{AdaFrame \cite{WuCVPR19}}                     & 80.2 \\
  \text{SCSampler \cite{KorbarICCV9}}                 & 81.0 \\
  \text{AR-Net \cite{Meng20}}               & 81.3 \\
  \text{SMART \cite{GowdaAAAI21}}                     & 82.1 \\
  \text{AR-Net (EfficientNet backbone) \cite{Meng20}} & 84.4 \\
    \text{ObjectGraphs} \cite{GkalelisCVPR2021} & 84.6 \\
    AdaFocusV2 \cite{WangCVPR22} & 85.0 \\
  \text{ViGAT (proposed; ResNet backbone)}    & 86.0 \\
  \text{ViGAT (proposed; ViT backbone)} & \textbf{88.1}
  \end{tabular}}
\end{center}
\label{tbl:ExpResFcvid}
\end{table}

\begin{table}[!t]
\caption{Performance comparison on MiniKinetics.}
\begin{center}
{
\begin{tabular}{llc}
       && top-1(\%) \\
\hline
\parbox[t]{1mm}{\multirow{8}{*}{\rotatebox[origin=l]{90}{MiniKinetics 130K}}} &
  \text{LiteEval \cite{WuNIPS19}}                     & 61.0 \\
  &\text{SCSampler}  \cite{KorbarICCV9}                 & 70.8 \\
  & \text{AR-Net} \cite{Meng20}         & 71.7 \\
  & \text{FrameExit}  \cite{GhodratiCVPR2021} & 73.3 \\
  & \text{AR-Net  (EfficientNet backbone) \cite{Meng20}} & 74.8 \\
  & \text{FrameExit  (EfficientNet backbone) \cite{GhodratiCVPR2021}} & 75.3 \\
  &     AdaFocusV2 \cite{WangCVPR22} & 75.4 \\
  & \text{ViGAT (proposed; ResNet backbone)} & 70.6 \\
  & \text{ViGAT (proposed; ViT backbone)} & \textbf{78.8} \\
  \hline
  \parbox[t]{1mm}{\multirow{14}{*}{\rotatebox[origin=l]{90}{MiniKinetics 85K}}} &
   TBN \cite{LiAAAI19} & 69.5  \\
   & BAT \cite{Chi_CVPR_2020} & 70.6 \\
   & MARS (3D ResNet backbone)  \cite{CrastoCVPR19} & 72.8  \\
   & Fast-S3D (Inception backbone )  \cite{XieECCV18} & 78.0 \\
   & ATFR (X3D-S backbone)\cite{FayyazCVPR21} & 78.0\\
   & ATFR (R(2+1)D backbone) \cite{FayyazCVPR21} & 78.2 \\
   & RMS (SlowOnly backbone) \cite{KimCVPR20} & 78.6 \\
   & ATFR (I3D backbone) \cite{FayyazCVPR21}&  78.8 \\
   & Ada3D  (I3D backbone on Kinetics) \cite{LiCVPR21}& 79.2 \\
   & ATFR (3D Resnet backbone) \cite{FayyazCVPR21} &  79.3 \\
   & CGNL (Modified ResNet backbone) \cite{Kaiyu_NIPS_2018} & 79.5 \\
   & TCPNet (ResNet backbone on Kinetics) \cite{GaoNIPS21} &  80.7 \\
   & LgNet (R3D Backbone) \cite{ZhouTMM2022}& 80.9 \\
   & ViGAT (proposed; ResNet backbone) & 74.3 \\
   & ViGAT (proposed; ViT backbone) & \textbf{82.1}
\end{tabular}}
\end{center}
\label{tbl:ExpResMiniKinetics}
\end{table}

\begin{table}[!t]
\caption{Performance comparison on ActivityNet.}
\begin{center}
{
\begin{tabular}{lc}
       & mAP(\%) \\
\hline
  \text{AdaFrame \cite{WuCVPR19}} & 71.5 \\
  \text{ListenToLook \cite{GaoCVPR2020}} & 72.3 \\
  \text{LiteEval \cite{WuNIPS19}} & 72.7 \\
  \text{SCSampler \cite{KorbarICCV9}} & 72.9 \\
  \text{AR-Net \cite{Meng20}} & 73.8 \\
  \text{FrameExit \cite{GhodratiCVPR2021}} & 77.3 \\ 
  AdaFocusV2 \cite{WangCVPR22} & 79.0 \\
  \text{AR-Net (EfficientNet backbone) \cite{Meng20}} & 79.7 \\
  \text{MARL (ResNet backbone on Kinetics) \cite{WuICCV19}} & 82.9 \\
  \text{FrameExit (X3D-S backbone) \cite{GhodratiCVPR2021}} & 87.4 \\
    \text{ViGAT (proposed; ResNet backbone)} & 82.1 \\
  \text{ViGAT (proposed; ViT backbone)} & \textbf{88.1}
  \end{tabular}}
\end{center}
\label{tbl:ExpResActnet}
\end{table}

The proposed approach is compared against the top-scoring approaches of the literature on the three employed datasets, specifically,  TBN \cite{LiAAAI19}, BAT \cite{Chi_CVPR_2020}, MARS \cite{CrastoCVPR19}, Fast-S3D \cite{XieECCV18}, RMS \cite{KimCVPR20}, CGNL \cite{Kaiyu_NIPS_2018},  ATFR  \cite{FayyazCVPR21}, Ada3D  \cite{LiCVPR21}, TCPNet \cite{GaoNIPS21}, LgNet \cite{ZhouTMM2022}, ST-VLAD \cite{SoltanianAG20}, PivotCorrNN \cite{SunghunECCV18}, LiteEval \cite{WuNIPS19}, AdaFrame \cite{WuCVPR19}, ListenToLook \cite{GaoCVPR2020}, SCSampler \cite{KorbarICCV9}, AR-Net \cite{Meng20}, SMART \cite{GowdaAAAI21}, ObjectGraphs \cite{GkalelisCVPR2021}, MARL \cite{WuICCV19}, FrameExit \cite{GhodratiCVPR2021} and AdaFocusV2 \cite{WangCVPR22} (note that not all of these works report results for all the datasets used in the present work).
The reported results on FCVID, MiniKinetics and ActivityNet are shown in Tables \ref{tbl:ExpResFcvid},  
\ref{tbl:ExpResMiniKinetics} and \ref{tbl:ExpResActnet}, respectively.
The majority of the methods utilize a ResNet-like backbone (sometimes pretrained on ImageNet) and train it from scratch (or fine-tune it) on the respective dataset; when this is not the case, in brackets next to the name of each method we denote the different backbone (e.g. EfficientNet, X3D, etc.) and/or the dataset used for training it (e.g. Kinetics).
From the obtained results we observe the following:

i) The proposed approach achieves the best performance in all datasets, improving the state-of-the-art by 3.1\%, 3.4\%, 1.2\% and 0.7\% on FCVID,  MiniKinetics 130K and 85K, and ActivityNet, respectively.
We should also note that the proposed model exhibits a very stable behavior converging to the above values, as shown in the plot of Fig. \ref{fig:converenge}.

ii) Concerning our ViGAT variant that utilizes a ResNet backbone pretrained on ImageNet, this outperforms the best-performing literature approaches that similarly use a ResNet backbone in FCVID and ActivityNet (see Tables  \ref{tbl:ExpResFcvid} and \ref{tbl:ExpResActnet}).
Specifically, we observe a significant performance gain of 1\% over AdaFocusV2 \cite{WangCVPR22}, which is the previous state-of-the-art method.
We also see that ViGAT provides a performance improvement of 1.4\% over ObjectGraphs \cite{GkalelisCVPR2021}, which is the best previous bottom-up method.
The above result clearly demonstrates the advantage of our architecture, i.e. the use of a pure-attention head in order to capture effectively both the spatial information and long-term dependencies within the video, instead of using an attention-LSTM structure as in \cite{GkalelisCVPR2021}.
We also observe a large gain of 3.1\% over AdaFocusV2 (the previous top-performing approach with ResNet backbone) on ActivityNet.
We should also note that in some cases ViGAT even with a ResNet backbone outperforms methods utilizing a stronger backbone, e.g. the AR-Net with the EfficientNet backbone on FCVID and ActivityNet \cite{Meng20}.
On the other hand, this is not the case in the MiniKinetics dataset.
This is attributed to the fact that our ImageNet-pretrained backbone is frozen, used as a feature extractor; whereas the above methods train or fine-tune the employed ResNet backbone in the larger MiniKinetics dataset, leading naturally to improved performance.

iii) The use of ViT instead of the ResNet backbone in ViGAT, i.e. the proposed pure-attention approach, provides a considerable performance boost: 2.1\% on FCVID, and an impressive 8.2\%, 7.8\% and 6\% on MiniKinetics 130K, 85K and ActivityNet.
The latter may be explained by the fact that ActivityNet and MiniKinetics contain a more heterogeneous mix of short- and long-term actions, and thus a stronger backbone that provides a better representation of the objects can facilitate the discrimination of a larger variety of action/event types.
This behavior has also been observed in other methods, e.g., AR-Net (using ResNet and EfficientNet) and FrameExit (using ResNet and X3D-S), as illustrated in Tables \ref{tbl:ExpResFcvid} and \ref{tbl:ExpResActnet}.

Concerning computational complexity, the Fvcore Flop Counter \cite{fvcorFlopCount} is used to compute the FLOPs (floating point operations) of the ViGAT head and ViT backbone.
For the Faster R-CNN object detector, due to its inherent randomness during the inference stage, we utilize the GFLOPs per frame reported in \cite{Carion_2020_ECCV}.
Using the above tool, we verified that the proposed ViGAT head is very lightweight, with 3.85 million parameters and only 3.87 GFLOPs to process a video in MiniKinetics.
On the other hand, counting also the execution of the Faster R-CNN \cite{renNips2015faster} object detector and the ViT backbone \cite{DosovitskiyICLR21} applied on each object and frame increases the total complexity of our method to 34.4 TFLOPs.
The latter figure is comparable with the complexity of some of the most recent top-down approaches of the literature, such as ViViT Large and Huge \cite{Arnab_2021_ICCV} with 11.9 and 47.7 TFLOPs, respectively.
However, we should note that during ViGAT training, the pre-trained Faster R-CNN and ViT backbone that are the most computationally expensive components of ViGAT are executed only once per video, yielding a dramatic GFLOP reduction for the overall training procedure.
Thus, compared to the video transformer models mentioned above, 
which were trained on dedicated high-performance tensor processing accelerators,
ViGAT has a significantly lower training complexity that allowed all reported experiments to run on single-GPU PCs.
Moreover, the overall complexity of ViGAT can be optimized by using more efficient pre-trained networks for object detection and feature representation, such as the ones presented in \cite{Li_WFIoT_2020,Zhang_NIPS_2021}, which report a considerably smaller number of GFLOPs than  \cite{renNips2015faster,DosovitskiyICLR21}.

\subsection{Event recognition ablation study}
\label{ssec:EventAblationStudy}

In order to gain a further understanding of the proposed event recognition approach, results of two ablation experiments are presented in this section.
These experiments are performed using the ViGAT with ViT backbone and following the training procedure described in Section \ref{ssec:Setup}.
Specifically, we perform:
\begin{itemize}
    \item Assessment of the impact of the weight sharing scheme, as well as the relative importance of the object and frame feature information, on the performance of our model.
    \item Investigation of the effect of using a different number of layers within the GAT blocks of the proposed architecture.
\end{itemize}

\begin{table}[!ht]
\begin{center}
\caption{Influence of the number of layers in the GAT blocks of ViGAT with ViT backbone along three datasets.}
\label{tab:ablationGcnDepths}
\begin{tabular}{l|cccc}
 Dataset \textbackslash \;  Number of layers  &  1 & 2 & 3 & 4 \\ \hline
 FCVID (mAP(\%)) &  \textbf{88.12} & 88.10 & 88.04 & 87.93 \\
 MiniKinetics 85K (top-1(\%))  &  81.58 & \textbf{82.16} & 81.22 & 80.76 \\
 ActivityNet (mAP(\%)) &  88.12 & 88.11 & \textbf{88.15} & 87.86
\end{tabular}
\end{center}
\end{table}

\begin{figure*}[!htb]
\centering
\includegraphics[width=1.99\columnwidth]{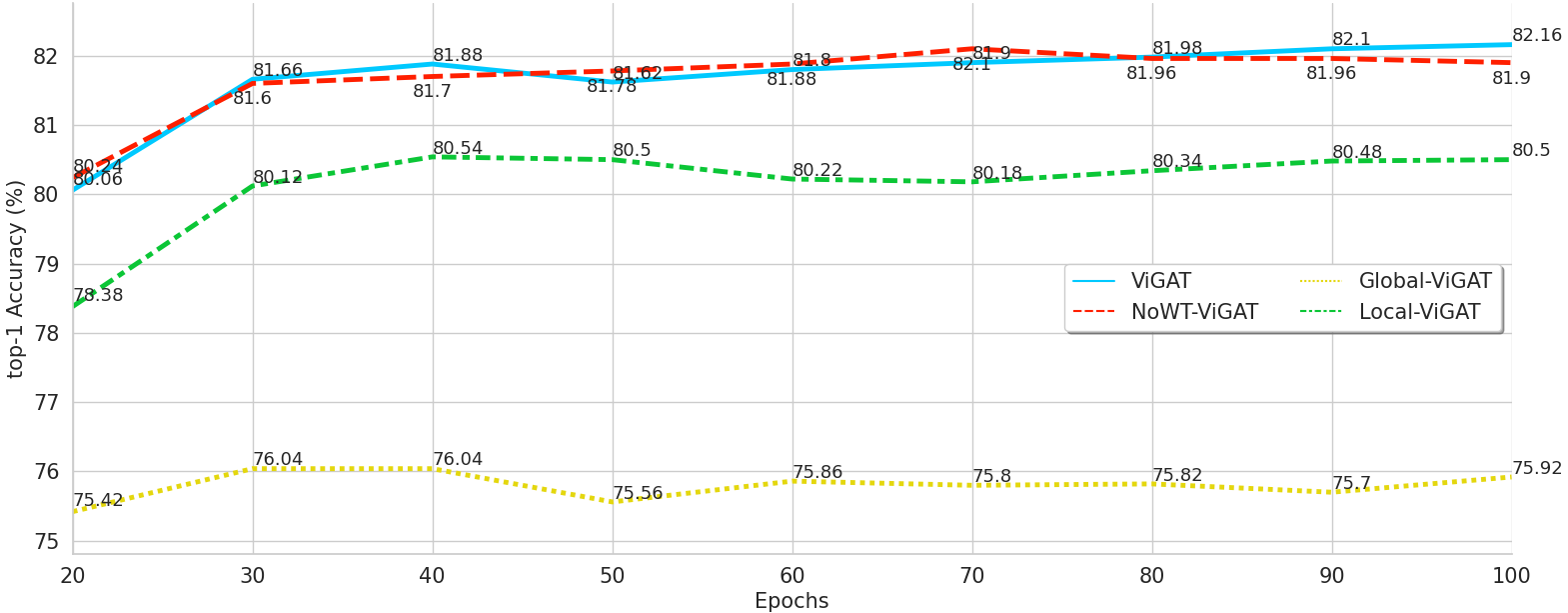}
\caption{Ablation study in MiniKinetics 85K, evaluating four variants of our model, i.e., ViGAT (proposed model with weight-tying applied), NoWT-ViGAT (proposed model without weight-tying), Global-ViGAT (model variant using only frame feature representations) and Local-ViGAT (model variant using only object feature representations). For each model variant, the top-1(\%) performance is plotted.
We see that the object features provide significant bottom-up information for the recognition of the video event, and that their combination with the global frame features leads to considerable performance gains.}
\label{fig:converenge}
\end{figure*}

In the first ablation experiment, we utilize MiniKinetics 85K to evaluate the performance of four different variants of our method: i) ViGAT: our proposed model (Section \ref{ssec:Results}), i.e. with weight-tying applied across the three GAT blocks, ii) noWT-ViGAT: this model has the same architecture as ViGAT with the difference that the weights are not shared along the three GAT blocks (i.e. the blocks $\Omega1$, $\Omega2$ and $\Omega3$ of Fig. \ref{fig:Architecture} have different weights), iii) Global-ViGAT: this model utilizes only the GAT block $\Omega1$ to process only the frame feature representations \eqref{E:feaMatGlobal}, iv) Local-ViGAT: contrarily to the above, this model employs only the GAT blocks $\Omega2$ and $\Omega3$, i.e. the branch of the ViGAT head that processes the object feature representations \eqref{E:feaMatLocal}.
The evaluation performance in terms of top-1(\%) for all models along the different epochs is shown in Fig. \ref{fig:converenge}.
From the obtained results we observe the following:

i) The Local-ViGAT model outperforms Global-ViGAT with a high absolute top-1(\%) gain of $4.58\%$, demonstrating the significance of the bottom-up information (represented by the object features) and the effectiveness of our approach in exploiting this information.
Moreover, we observe that the object and frame features are to some extent complementary, as shown by the $1.66\%$ absolute top-1(\%) performance gain of ViGAT (which exploits both features) over the Local-ViGAT.

ii) ViGAT outperforms NoWT-ViGAT in MiniKinetics 85K by 0.26\% absolute top-1(\%), showing that  the use of shared weights along the different GAT blocks may act as a form of regularization stabilizing the training procedure, as for instance has been observed in \cite{InanICLR2017,Lan_2020_ICLR,GuohaoICML2021}.
However, we should note that this is not necessarily always the case, i.e. for other datasets a larger network capacity may be beneficial. Besides potentially improved event recognition results, the use of shared weights leads to reduced memory footprint: using the Fvcore Flop Counter \cite{fvcorFlopCount} we can see that NoWT-ViGAT has 8.426 million parameters. In comparison, the proposed ViGAT (3.85 million parameters) achieves a $2.3\times$ lower memory footprint. 

In a second ablation experiment, the influence of the number of GAT layers $M$ \eqref{E:DgcnLayer} in the performance of ViGAT is examined.
Specifically, $M$ within each block (Fig. \ref{fig:Architecture}) is varied from 1 to 4 and the performance is recorded.
From the results shown in Table \ref{tab:ablationGcnDepths},
we observe that $M=2$ is optimal or nearly optimal along all three datasets (for simplicity, concerning MiniKinetics we run this ablation experiment only on its 85K variant), and the performance starts to decrease for $M > 3$.
This behaviour has been often observed in the literature and is attributed to the well-known oversmoothing problem of GNNs \cite{Chen2020}.

\subsection{Event explanation results and ablation study}
\label{ssec:EvExplanationResults}

In this section, the proposed explainability approach (Section \ref{sec:MultiExplanations}) with the ViT backbone is evaluated on the ActivityNet dataset.
This dataset is selected here because its videos are represented with a large number of frames (i.e. $N = 120$), allowing for a thorough evaluation of different XAI criteria.

Firstly, we perform a quantitative evaluation using the XAI measures described in Section \ref{ssec:Measures}.
Specifically, the various criteria are evaluated based on their ability to select the $\Upsilon$ most salient frames explaining model's outcome, where $\Upsilon$ is set to $\Upsilon= 1, 2, 3, 5, 10$ and $20$.

\begin{figure}[!ht]
\centering
\includegraphics[width=.99\columnwidth]{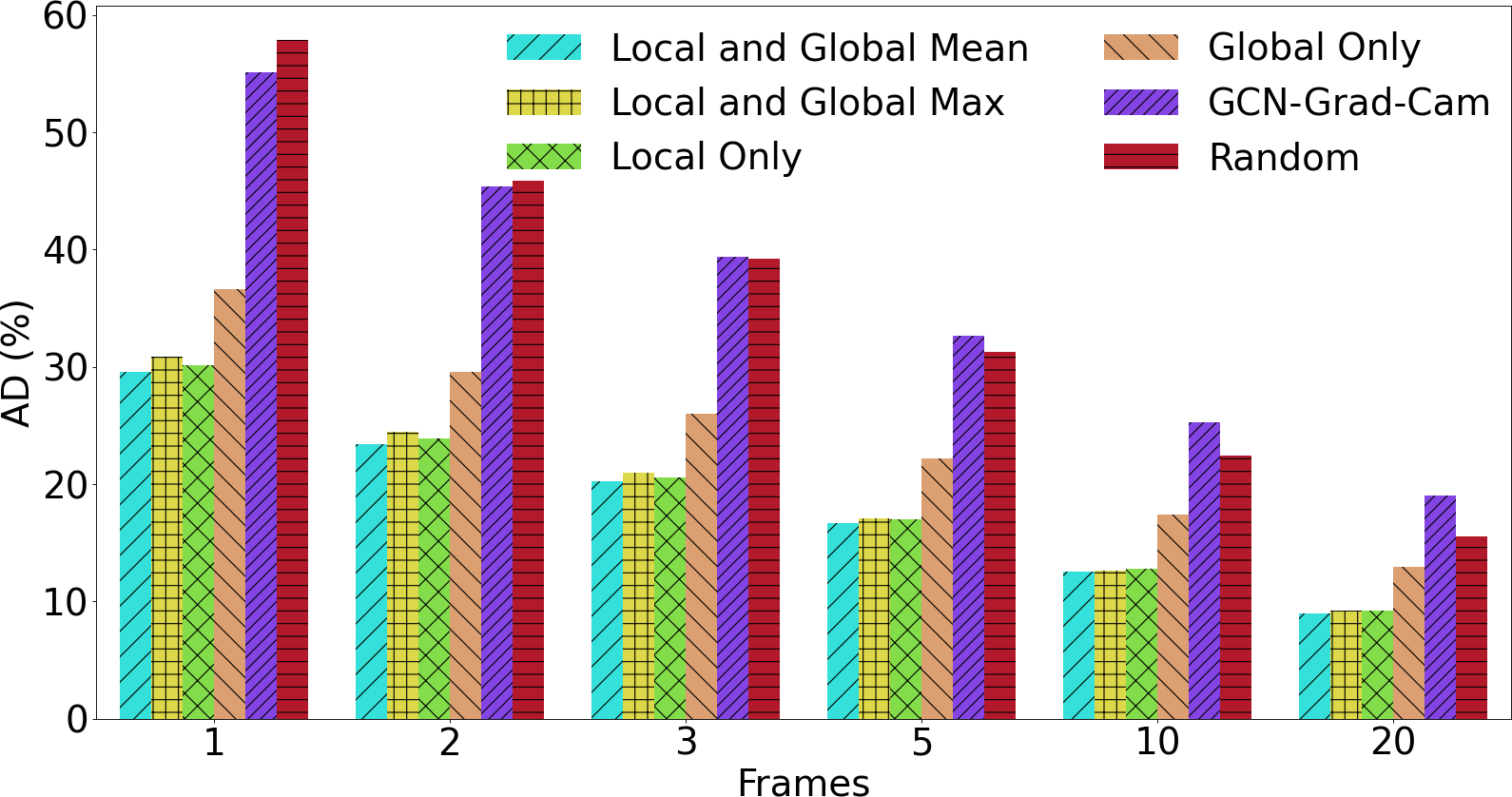}
\caption{Average drop ($AD$) performance along varying number of frames for the six considered XAI criteria.
Lower values are better.}
\label{fig:averagedrop}
\end{figure}

\begin{figure}[!ht]
\centering
\includegraphics[width=.99\columnwidth]{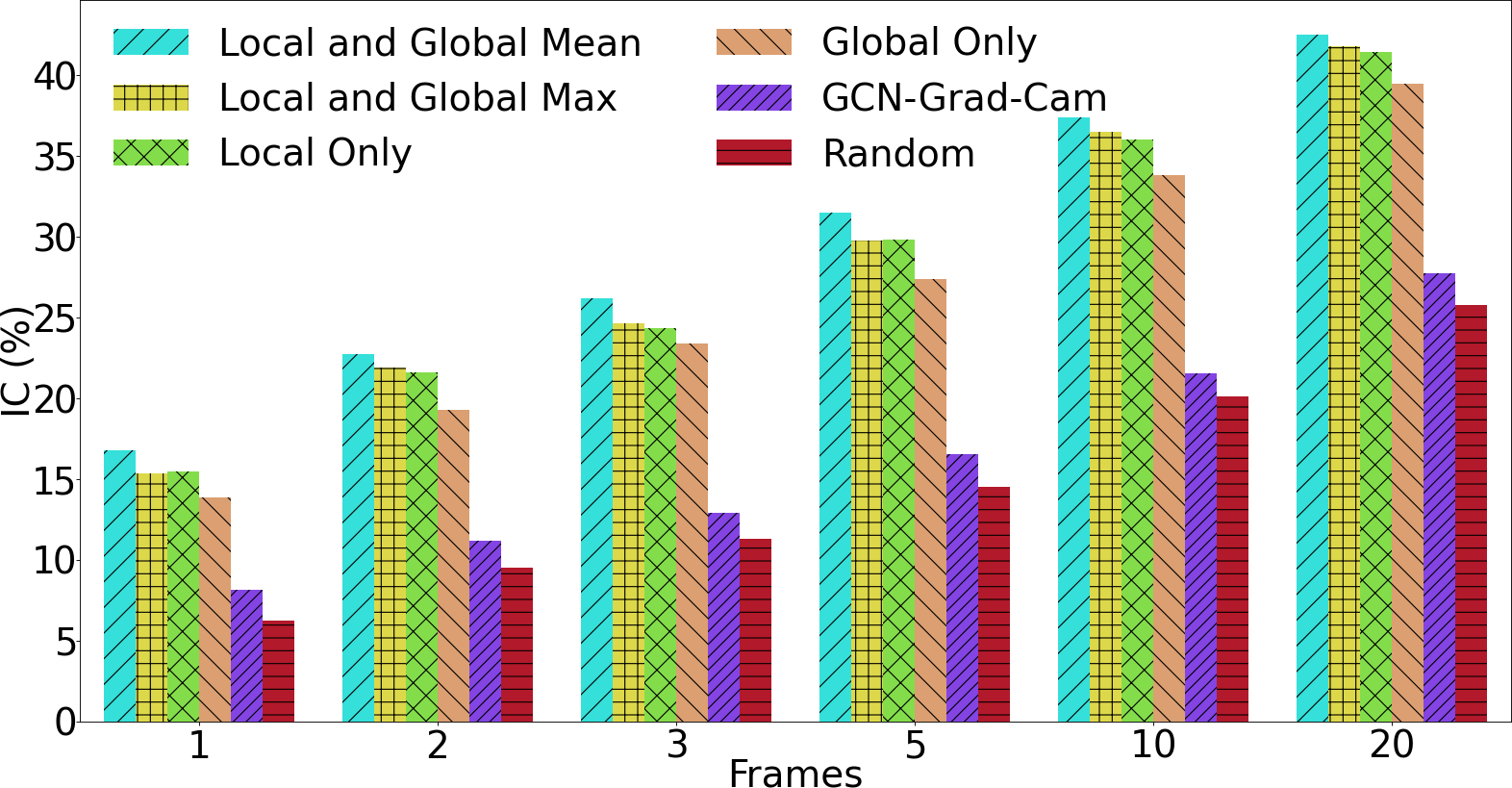}
\caption{Increase in confidence ($IC$) performance along varying number of frames for the six considered XAI criteria.
Higher values are better.}
\label{fig:ncconf}
\end{figure}

\begin{figure}[!ht]
\centering
\includegraphics[width=.99\columnwidth]{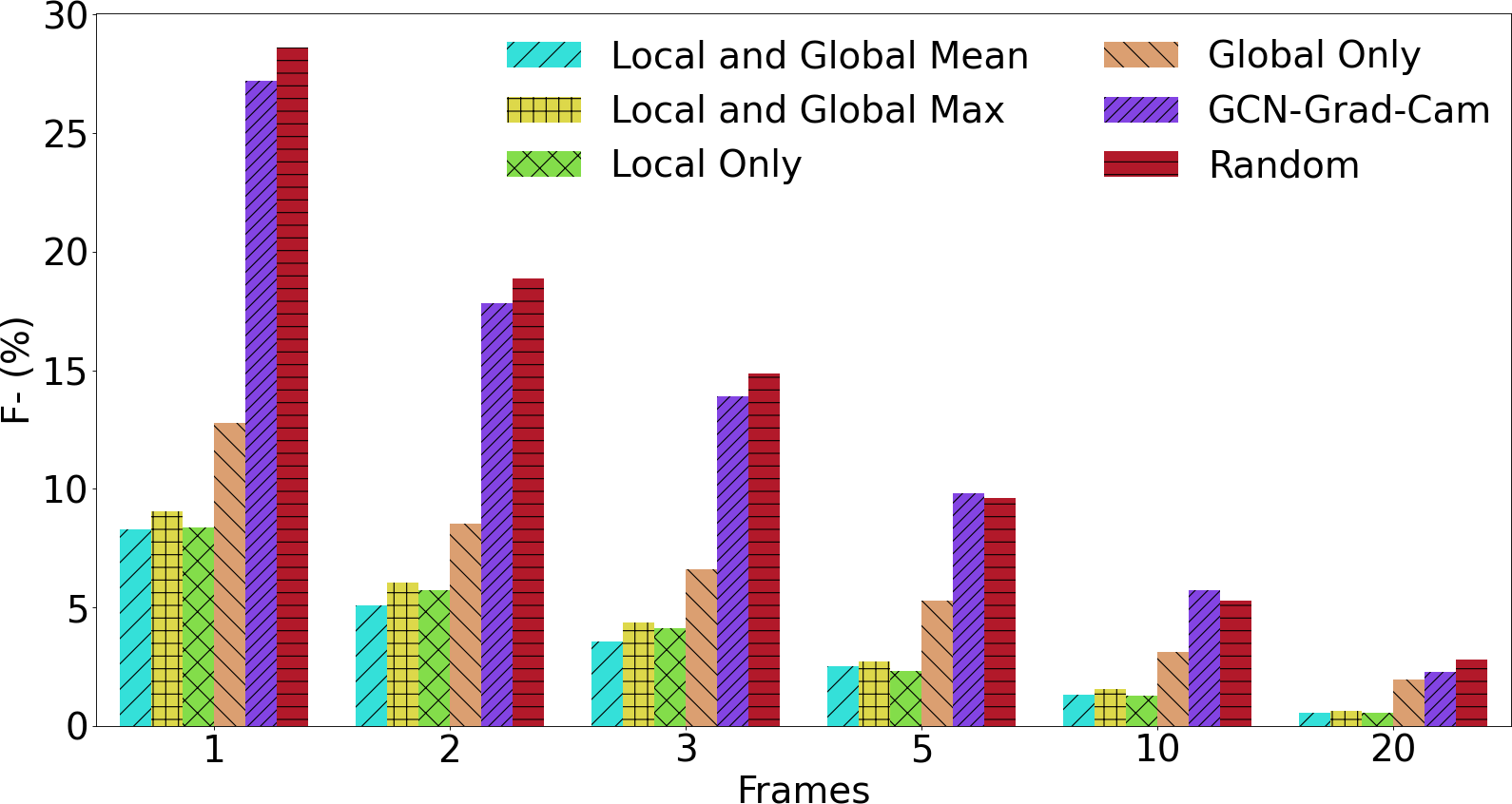}
\caption{Fidelity minus ($F-$) performance along varying number of frames for the six considered XAI criteria.
Lower values are better.}
\label{fig:fidelityminus}
\end{figure}

\begin{figure}[!ht]
\centering
\includegraphics[width=.99\columnwidth]{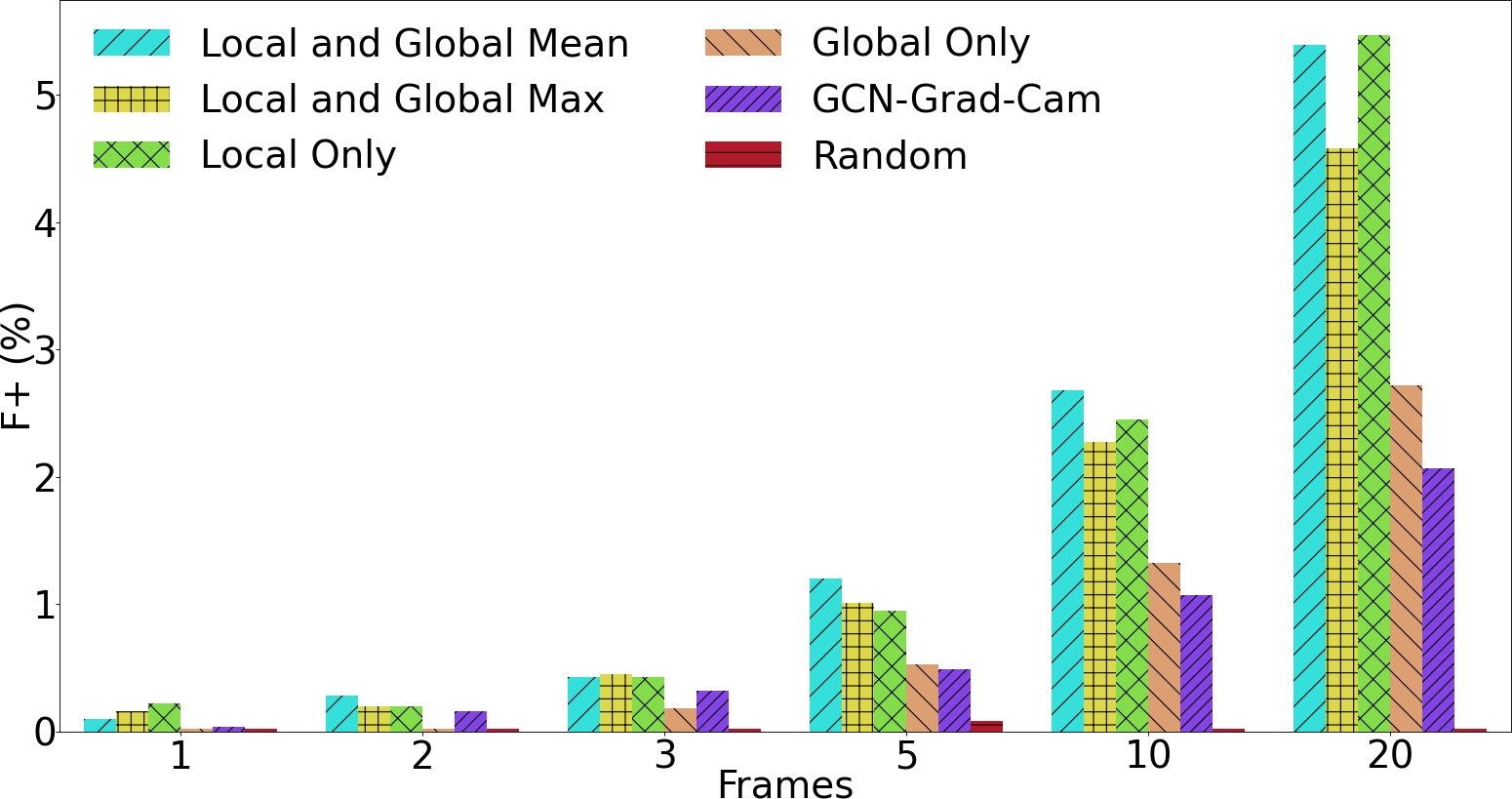}
\caption{Fidelity plus ($F+$) performance along varying number of frames for the six considered XAI criteria.
Higher values are better.}
\label{fig:fidelityplus}
\end{figure}

\begin{figure*}[!htb]
\begin{center}
\includegraphics[width=0.999\linewidth,height=.18\linewidth]{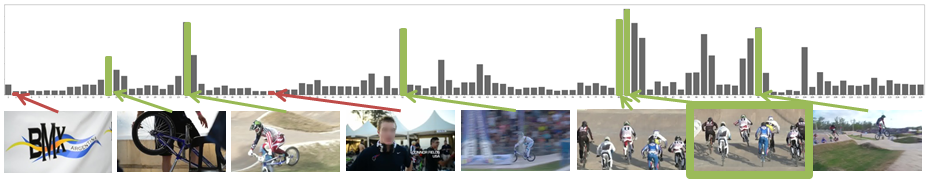}
\end{center}
\caption{Explanation example for a video correctly categorized into class ``BMX''.
The barplot of frame $\beta^{(n)}$ values \eqref{E:widFrameMean} is provided at the top of the figure.
The two and six video frames with lowest and highest $\beta^{(n)}$ (depicted with red and green bars, respectively) are shown below the barplot.
The video frame corresponding to the highest $\beta^{(n)}$ is placed within a green rectangle.
We see that the model focuses on the frames that contain at least one bike and ignores other irrelevant ones (e.g. the computer graphics frame, appearing first from the left in the figure).
It is also worth noting that the frame selected as the most salient (i.e., with highest $\beta^{(n)}$) is the one that depicts multiple BMX vehicles.}
\label{fig:bmx}
\end{figure*}

\begin{figure*}[!htb]
\begin{center}
\includegraphics[width=0.999\linewidth,height=.18\linewidth]{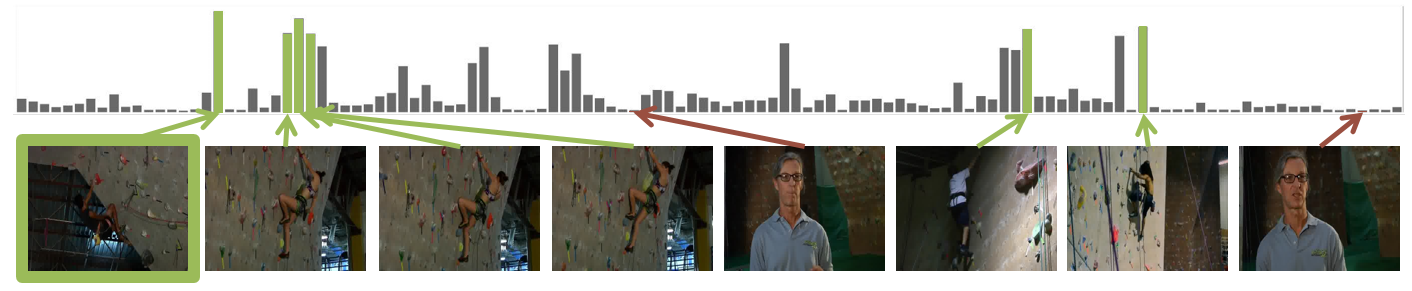}
\end{center}
\caption{Explanation example for a video correctly categorized into the class “Rock climbing”. Based on the $\beta^{(n)}$ values \eqref{E:widFrameMean} we see that the classifier focuses on the frames showing a wall and a climber to classify this video, while frames irrelevant to the underlying event (e.g. the frames depicting an interview) receive a low $\beta^{(n)}$ value and are thus disregarded.}
\label{fig:climbing}
\end{figure*}

\begin{figure*}[!htb]
\begin{center}
\includegraphics[width=0.999\linewidth,height=.18\linewidth]{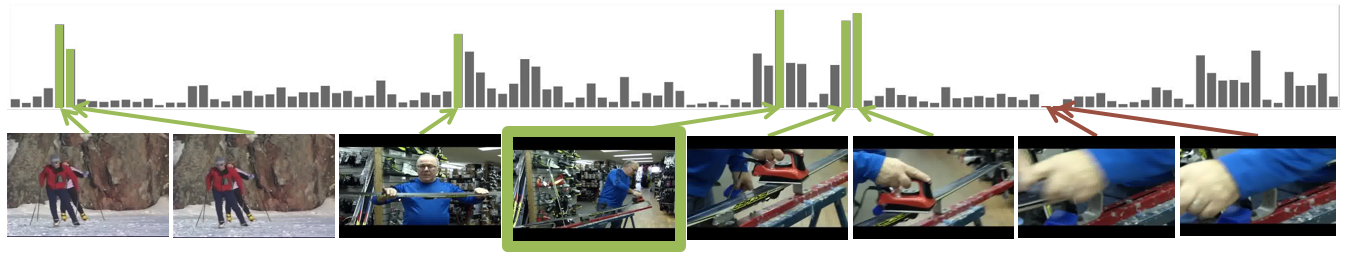}
\end{center}
\caption{Explanation example for a video correctly categorized into the class “Waxing Skis”. This is a hard example because, as we can see from the two left-most frames in this figure, frames showing a skier and snow are part of the video and are even assigned high $\beta^{(n)}$ values \eqref{E:widFrameMean}; these could mislead to classifying the video as "Skiing" (which is among the events included in this dataset). However, thanks to the highest $\beta^{(n)}$ values being assigned by the proposed ViGAT to frames that depict waxing skis, the classifier correctly recognizes this event.}
\label{fig:waxing}
\end{figure*}

\begin{figure*}[!htb]
\begin{center}
\includegraphics[width=0.999\linewidth,height=.18\linewidth]{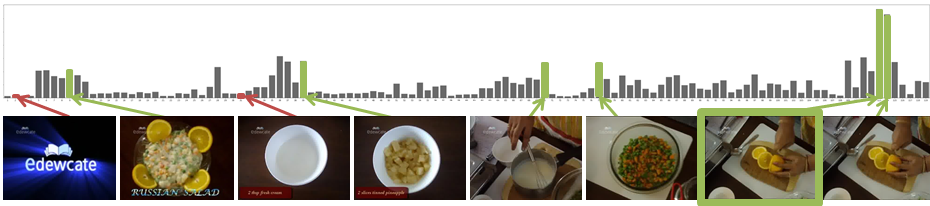}
\end{center}
\caption{Explanation example for a video belonging to class ``Preparing salad'' but miscategorized as ``Making lemonade''.
As with previous examples, the $\beta^{(n)}$ values correctly indicate the frames that are irrelevant to the recognized class, e.g. the two frames with the lowest $\beta^{(n)}$ depict a computer graphics image and an empty bowl, respectively.
On the other hand, the two frames with highest $\beta^{(n)}$ show human hands cutting lemons, thus providing a convincing explanation why this video was misrecognized as ``Making lemonade'' by the proposed model.}
\label{fig:lemonade}
\end{figure*}

\begin{figure*}[!htb]
\begin{center}
\includegraphics[width=0.999\linewidth,height=.18\linewidth]{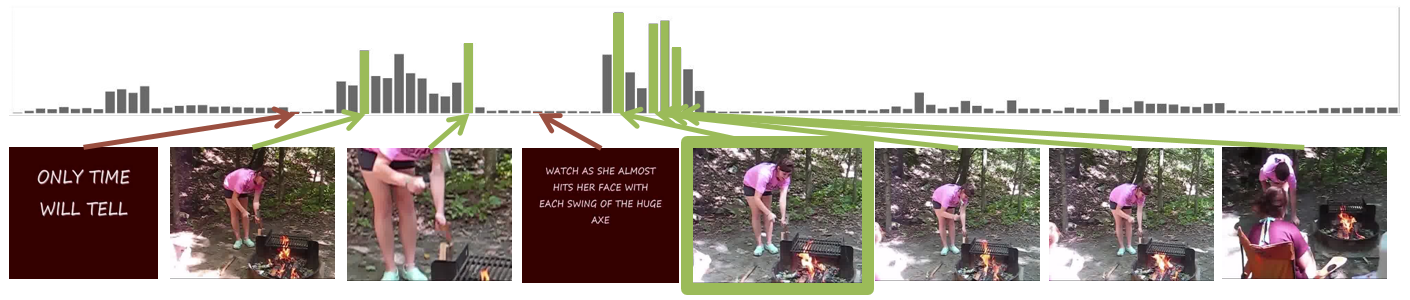}
\end{center}
\caption{Explanation example for a video belonging to class “Chopping wood” but miscategorized as “Starting a campfire”. 
The most salient frames (based on the $\beta^{(n)}$ values \eqref{E:widFrameMean}) are the ones depicting a person chopping wood next to a campfire. These frames provide a convincing explanation why the classifier has mistakenly labeled this video. On the other hand, we see that the most irrelevant frames to the classification decision (receiving a $\beta^{(n)}$ value close to zero) are the ones with overlay text on black frames. This video has many such frames, yielding a barplot that looks quite different from that of other videos.}
\label{fig:ChoppingWood}
\end{figure*}

\begin{figure*}[!htb]
\begin{center}
\includegraphics[width=0.999\linewidth,height=.18\linewidth]{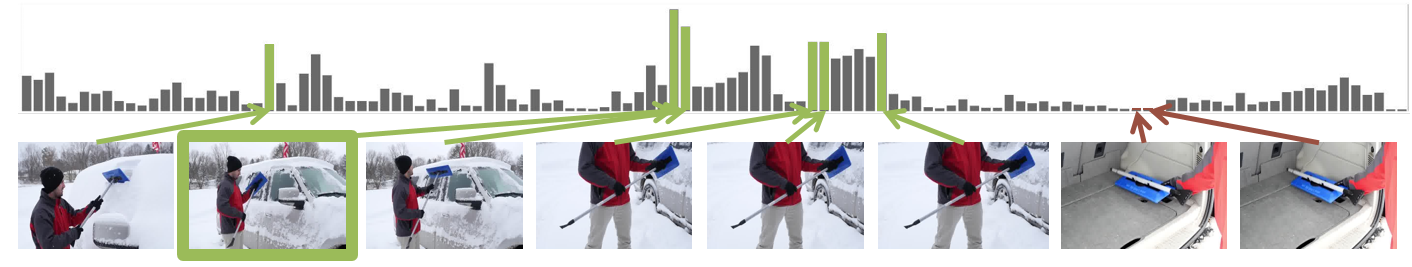} \\
\vspace{-3mm}
\hrulefill \\
\vspace{2mm}
\includegraphics[width=.3\linewidth,height=.21\linewidth]{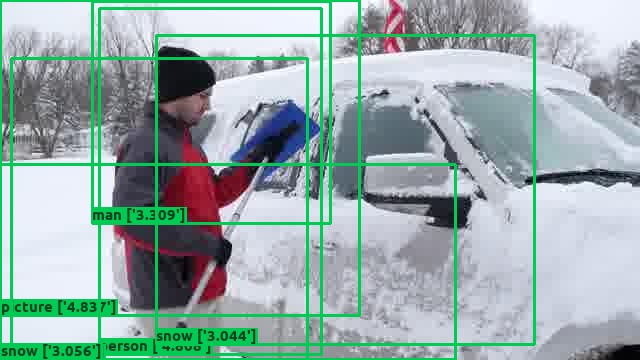} 
\includegraphics[width=.3\linewidth,height=.21\linewidth]{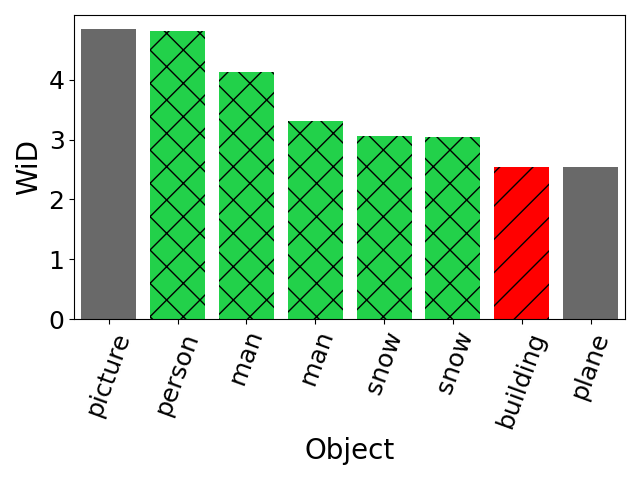} 
\includegraphics[width=.3\linewidth,height=.21\linewidth]{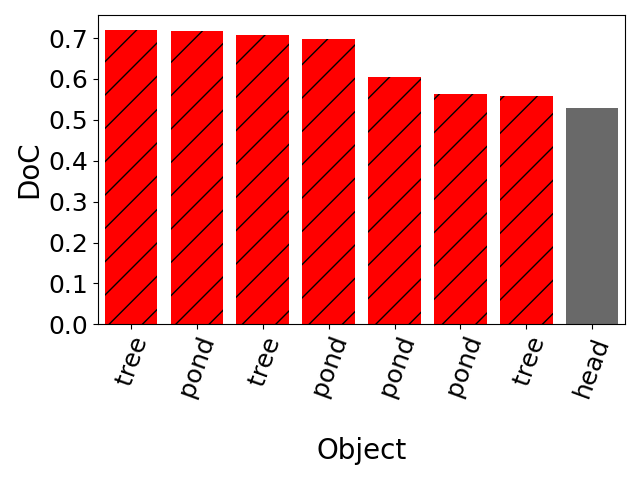} \\
\end{center}
\caption{Explanation example for a video belonging to class “Removing ice from car” but miscategorized as “Shoveling snow”.
We observe that the most salient frame (based on the $\beta^{(n)}$ values \eqref{E:widFrameMean}) depict a person removing ice from car, thus not providing enough evidence why this frame has been misclassified.
To this end, we resort to the object-level explanations, provided at the second row of the figure.
Specifically, the eight most salient objects are depicted for the most salient frame of the video, at the left side of the row, and the respective WiD values ($\omega_2^{(l,n)}$ \eqref{E:ObjectWiD}) are shown in the barplot at the middle of the row.
We also show a respective object detection barplot at the right side of the row, depicting the eight objects detected with the highest degree of confidence (DoC) value.
Concerning the bar colors: a green bar in the WiDs barplot indicates that the corresponding object did not appear in the top-8 DoC list but was promoted by our approach; 
a red bar indicates that this object is completely irrelevant with the recognized event.
We see that the most salient objects identified by our approach, i.e. ``person'', ``snow'', ``man'', etc., are not characteristic enough to differentiate between the two classes.
However, we observe that the object car, which is a differentiating factor between the two classes, is not detected by the object detector; additionally, the frame regions that the classifier focuses on do not include the car region, convincingly explaining the network's recognition decision in this failure example.}
\label{fig:shovellingSnow}
\end{figure*}

\begin{figure*}[!htb]
\begin{center}
\begin{tabular}{ccc}
\includegraphics[width=.3\linewidth,height=.21\linewidth]{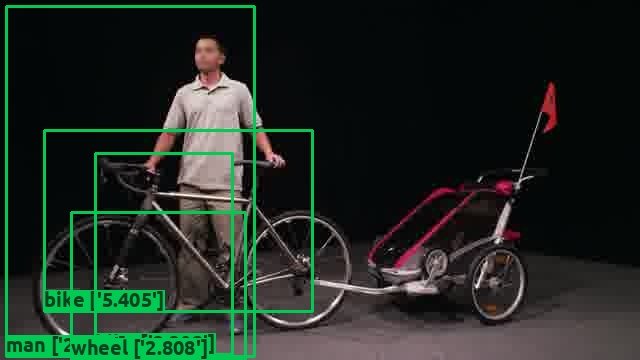} &
\includegraphics[width=.3\linewidth,height=.21\linewidth]{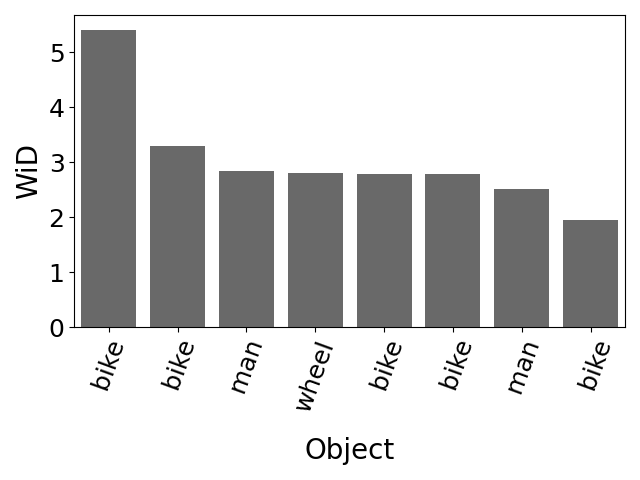} &
\includegraphics[width=.3\linewidth,height=.21\linewidth]{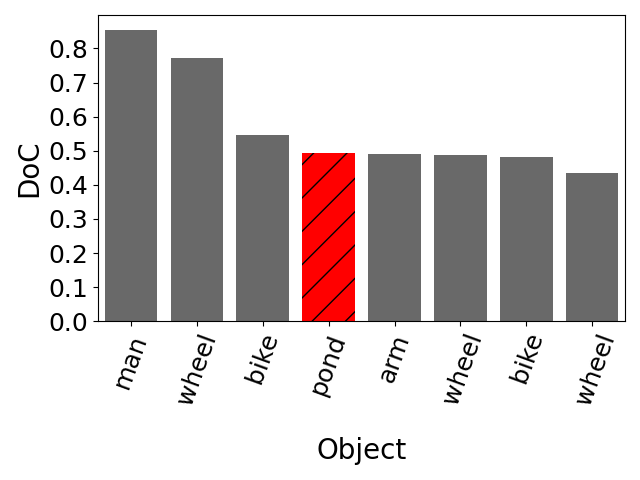} \\
\includegraphics[width=.3\linewidth,height=.21\linewidth]{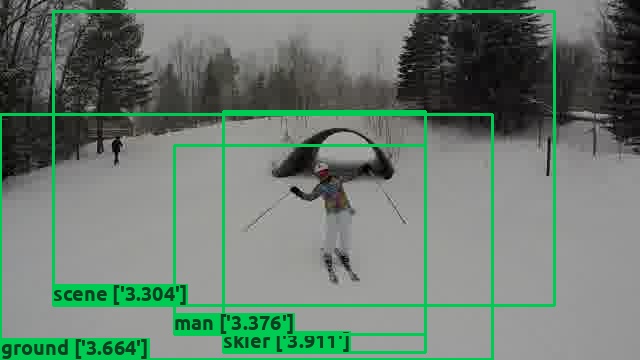} &
\includegraphics[width=.3\linewidth,height=.21\linewidth]{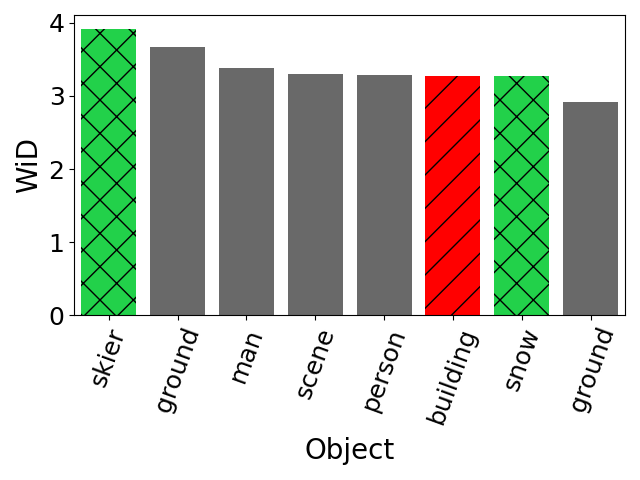} &
\includegraphics[width=.3\linewidth,height=.21\linewidth]{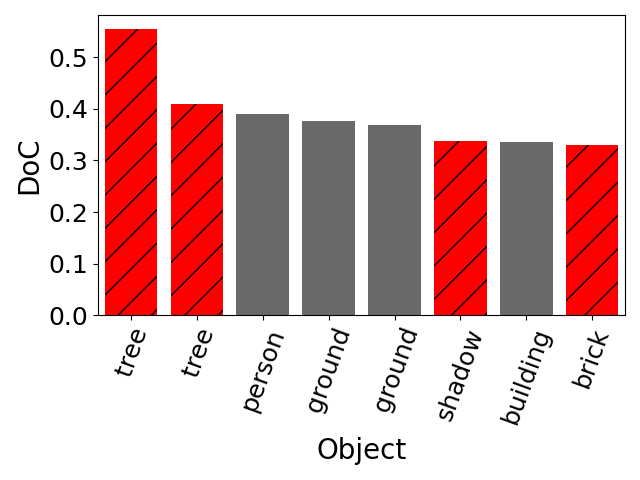} \\
\includegraphics[width=.3\linewidth,height=.21\linewidth]{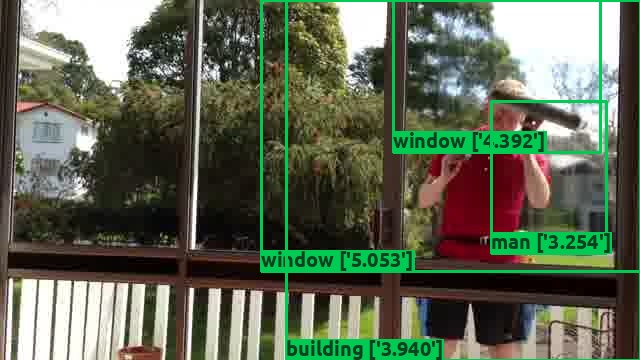} &
\includegraphics[width=.3\linewidth,height=.21\linewidth]{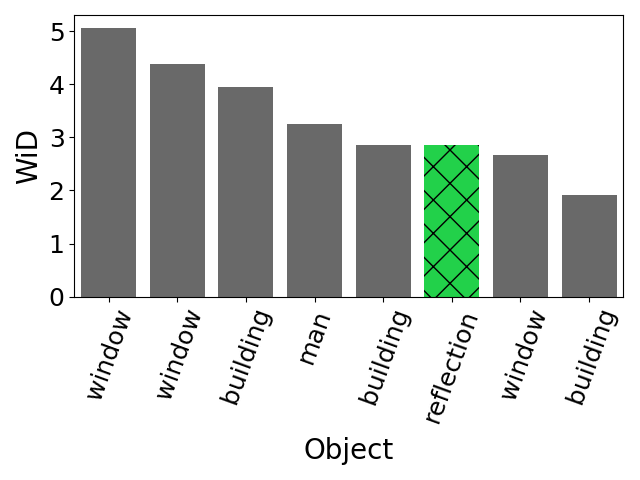} &
\includegraphics[width=.3\linewidth,height=.21\linewidth]{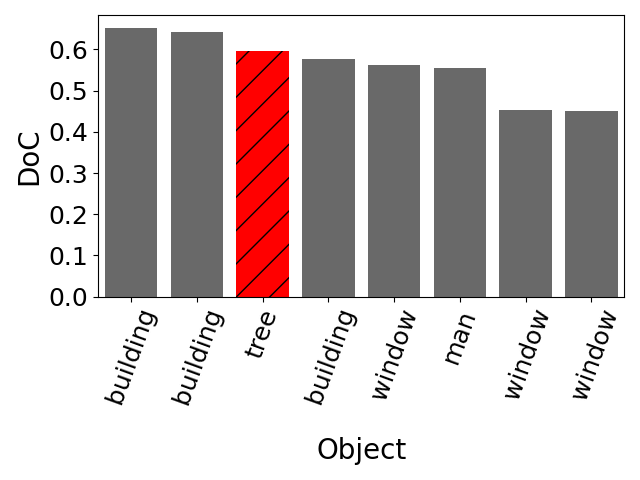} \\
\includegraphics[width=.3\linewidth,height=.21\linewidth]{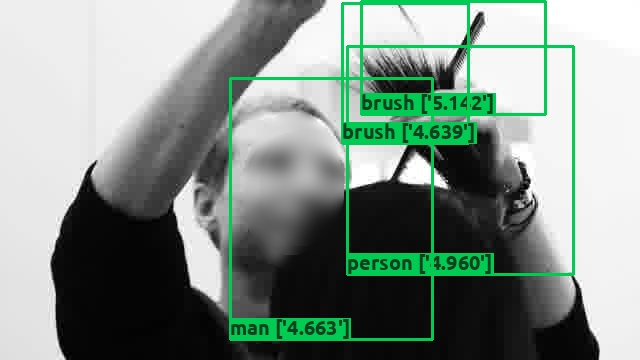} &
\includegraphics[width=.3\linewidth,height=.21\linewidth]{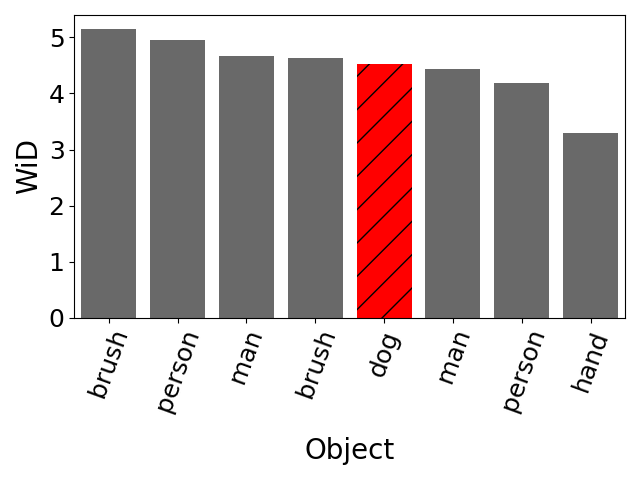} &
\includegraphics[width=.3\linewidth,height=.21\linewidth]{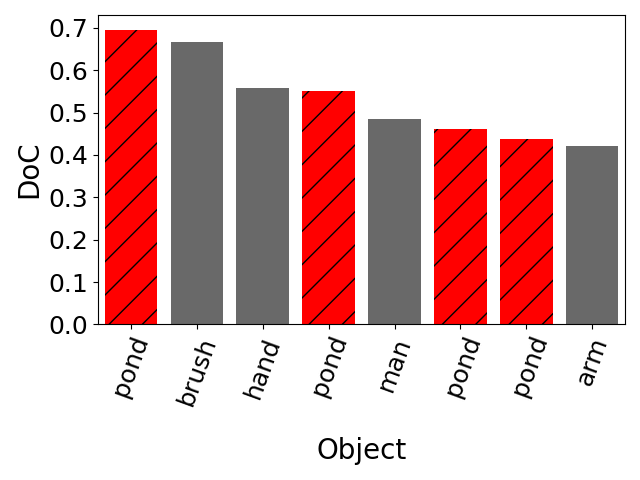} \\
\includegraphics[height=.24\linewidth]{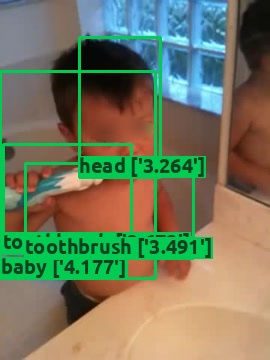} &
 \includegraphics[width=.3\linewidth,height=.24\linewidth]{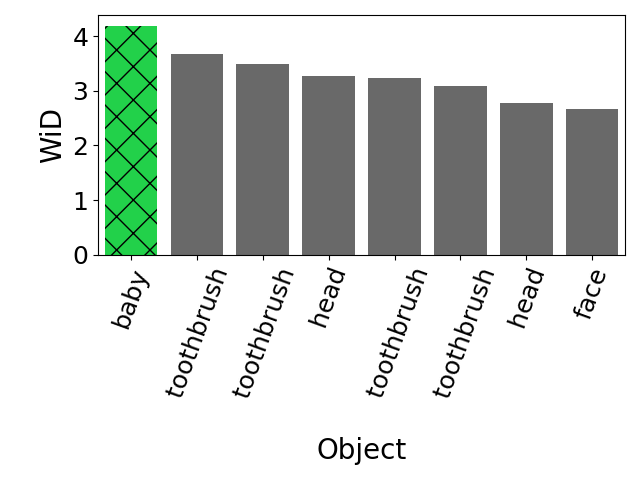} &
 \includegraphics[width=.3\linewidth,height=.24\linewidth]{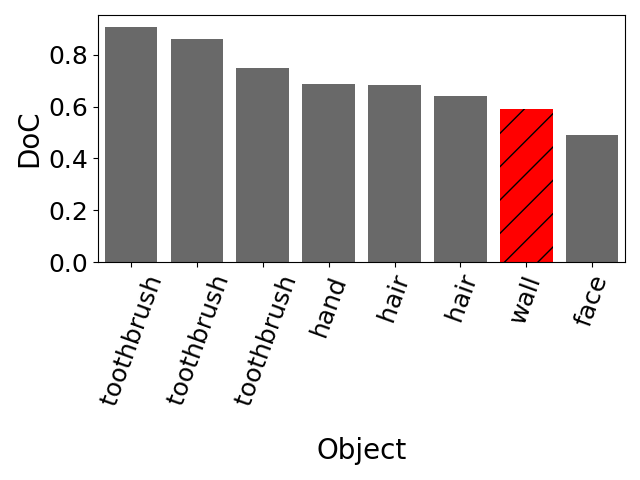}
\end{tabular}
\end{center}
\caption{Each row of this figure provides an explanation example produced using our approach for a video belonging to a different event category (from top to bottom):
a) ``Assembling a bike'', b) ``Skiing'',
c) ``Cleaning windows'', d) ``Getting a haircut'',
 e) ``Brushing teeth''.
An explanation example consists of the video frame associated with the highest $\beta^{(n)}$ (frame-level WiDs) and the four objects in this frame corresponding to the highest object-based WiDs.
The two barplots in the middle and right of each row depict the objects in the frame corresponding to the eight highest WiD or degree of confidence (DoC) values, respectively.
A green bar in the WiDs barplot indicates that the corresponding object did not appear in the top-8 DoC list but was promoted by our approach and convincingly explains the network's recognition decision, e.g. see the ``skier'' and ``baby'' objects in the examples of the second and fifth row.
On the other hand, a red bar in the barplots indicates that this object is completely irrelevant with the recognized event, e.g., see the ``tree'' objects in the examples of second and third row.
We observe that in most cases our approach indicates objects very relevant to the recognized event as explanations for the event recognition result (``dog'' in the fourth example is a notable exception).
In contrary, objects with high DoC, although may indeed be depicted in the frame, are often not related to the event recognized by the model and are correctly not considered by our WiD-based approach as good explanations.}
\label{fig:vidExplanation}
\end{figure*}

We assess the following four ViGAT-based criteria (which can be also considered as a form of ablation study examining the explanation power of the various WiD-based criteria of ViGAT): i) Local and Global Mean, i.e. the mean of the frame-level WiDs, $\beta^{(n)}$  \eqref{E:widFrameMean}; this is our proposed XAI criterion,
ii) Local and Global Max, i.e. $max (\omega_1^{(n)}, \omega_3^{(n)})$, 
iii) Local Only, $\omega_3^{(n)}$ \eqref{E:WiDLocal}, and, iv) Global Only, $\omega_1^{(n)}$  \eqref{E:WiDGlobal}.
Additionally, the above criteria are compared against i) GCN-Grad-Cam \cite{PopeCVPR19}, which is the closest approach to ours and can be applied to the ViGAT architecture, and ii) random frame selection, as a baseline.
For the latter (denoted hereafter simply as Random), random selection is repeated five times and the average is reported for each individual XAI measure.

The evaluation results in terms of $AD$ \eqref{E:averagedrop}, $IC$ \eqref{E:increaseinconf}, $F-$ \eqref{E:fidelityminus} and $F+$ \eqref{E:fidelityplus} are depicted in Figs. \ref{fig:averagedrop},
\ref{fig:ncconf},
\ref{fig:fidelityminus} and
\ref{fig:fidelityplus}, respectively.
From the obtained results we observe the following:

i) In all cases the proposed WiD-based XAI criteria outperform by a large margin the random frame selection.
Therefore, it is clear that the WiDs derived by the learned adjacency matrices 
in the proposed ViGAT architecture can provide valuable information for explaining the model's decision.

ii) The proposed criteria also outperform GCN-Grad-Cam across all performance measures.
For instance, for $\Upsilon = 1$ (i.e. when the single salient frame is considered) our proposed XAI criterion ($\beta^{(n)}$) provides an absolute explanation performance improvement of approximately 25\%, 9\% and 18\% over GCN-Grad-Cam in terms of $AD$, $IC$ and $F-$, respectively.

iii) The local WiDs are powerful explainability indicators, outperforming the global ones; this further highlights that bottom-up (i.e. object) information is crucial for the recognition of events in video.

iv) The combination of the local and global WiDs (using either operator) in most cases offers a small but noticeable performance gain, showing that these indicators are to some degree complementary.
For instance, we observe in Fig. \ref{fig:ncconf} that the mean WiDs provide consistently an absolute 2\% $IC$ performance gain over using any of the individual WiD indicators alone.

v) Generally, in terms of $AD$, $IC$ and $F-$, GCN-Grad-Cam exhibits a performance close to the random baseline.
In contrary, it achieves a much better $F+$ performance from the random baseline, as shown in Fig. \ref{fig:fidelityplus}.
This is in agreement with similar results in the literature, e.g. in \cite{PopeCVPR19}.
More specifically, we note that the computation of $AD$, $IC$ and $F-$ is based on the selection of the $\Upsilon$ most salient frames, while in contrary, $F+$ on the remaining $Q - \Upsilon$ least salient ones.
Based on this observation, we can say that $AD$, $IC$ and $F-$ correspond to the notion of sparsity (measure of localization of an explanation in a small subset of the graph nodes) and $F+$ resembles the notion of fidelity (measure of the decrease in classification accuracy when the most salient graph nodes are occluded), as sparsity and fidelity are defined in \cite{PopeCVPR19}.
In the experimental evaluation of the above work it is shown that GCN-Grad-Cam provides explanations of high fidelity but poor sparsity, similarly to the results obtained here.

In order to gain further insight into the proposed explainability approach, qualitative results (examples) are also given in Figs. \ref{fig:bmx} to \ref{fig:vidExplanation}.
In Figs. \ref{fig:bmx}, \ref{fig:climbing} and \ref{fig:waxing}, we show the six most salient and the two least salient frames selected using our explainability criterion $\beta^{(n)}$ from correctly-recognized videos belonging to class ``BMX'', ``Rock Climbing'' and ``Waxing Skis'', respectively.
In Fig. \ref{fig:bmx}, we see that all selected frames contain at least one BMX bike, while the one with the highest $\beta^{(n)}$ contains several bike instances.
Similarly, in Fig. \ref{fig:climbing} the climber and the climbing wall, and in Fig. \ref{fig:waxing} instances of a person waxing skis, are clearly shown in the selected frames.
Regarding Fig. \ref{fig:waxing}, despite this video being a difficult example due to containing instances of two related events (``Waxing Skis'' and ``Skiing''),  the classifier correctly gives more attention to the frames related to the actual, ``Waxing Skis'', event, rather than to the ones depicting the skiers and snow, thus achieving to correctly classify the video.
On the other hand, in all figures, the frames assigned a low $\beta^{(n)}$ depict information irrelevant to the recognized event and thus are correctly dismissed as potential explanations by our approach.

Contrarily to the above examples, Figs. \ref{fig:lemonade}, \ref{fig:ChoppingWood} and  \ref{fig:shovellingSnow} show failure cases: videos of the classes ``Preparing salad'', ``Chopping wood'' and ``Removing ice from car'' that have been miscategorized as ``Making lemonade'', ``Starting a campfire'' and ``Shoveling snow'' respectively.
As in the previous examples, we observe that the frames associated with the lowest $\beta^{(n)}$ are visually irrelevant to the recognized events and thus were correctly dismissed.
On the other hand, most of the frames associated with high $\beta^{(n)}$ as well as the ones corresponding to the top $\beta^{(n)}$ value, contain objects relevant to the recognized class, explaining why the classifier mislabelled these videos.
For instance, the most salient frames of the videos in Figs. \ref{fig:lemonade} and \ref{fig:ChoppingWood} depict pieces of lemons and a campfire, thus providing an explanation why the classifier misclassified these videos as ``Making lemonade'' and ``Starting a campfire'', respectively.
Similarly, in Fig. \ref{fig:shovellingSnow}, utilizing the object-level explanations for the most salient frame of the video (provided at the second row of this figure), we discover that the object car is not detected by the object detector, misleading ViGAT to miscategorize this video as ``Shoveling snow''.

Finally, Fig. \ref{fig:vidExplanation} depicts several examples of the frame- and object-level explanations generated by our model: in each row, the selected best video frame explanation, as well as the top four 
object-explanations within each frame, as identified by our approach, are shown.
Additionally, two barplots per row are provided, depicting the eight objects with the highest WiD ($\omega_2^{(l,n)}$, see \eqref{E:ObjectWiD}) and degree of confidence values (the latter being an output of the employed object detector), respectively.
The same type of information is also been provided in the second row of Fig. \ref{fig:shovellingSnow} to help us understand why ViGAT misclassified that video.
We observe that the objects associated with the highest WiDs are well correlated with the recognized event.
Moreover, in most cases (i.e. when the object detector provides a correct object class detection) the class names of the objects can be used to provide a sensible semantic recounting \cite{GkalelisICIP2014} that describes the event detected in the video in a human-comprehensible format.
On the other hand, the same cannot be said for the objects associated with high degree of confidence values; these provide a general overview of the various objects depicted in the frame, rather than an insight on which of the depicted objects led to the event recognition decision.

\subsection{Limitations}
\label{sec:Limitations}

As shown from the experimental results, due to the extraction of bottom-up information and the utilization of attention at various levels of ViGAT, our method attains improved event recognition performance and has the ability to provide comprehensive explanations about the decision of the classifier.
However, as expected, in comparison to efficient top-down approaches, the above achievements come with a high cost in memory consumption and inference time.
To this end, we have tied the weights of the three GAT blocks of ViGAT, achieving more than $2 \times$ improvement in memory utilization (see Section \ref{ssec:EventAblationStudy}).  
However, the computational overhead is mainly due to the use of the object detector at each sampled frame, to extract a set of objects, and the subsequent use of a backbone network (ViT) to provide a feature representation of them (i.e. to derive the bottom-up information).
To reduce this overhead, inspired by the relevant literature \cite{Meng20,GowdaAAAI21,GhodratiCVPR2021}, we plan on investigating techniques for selecting only a small fraction of the sampled frames to use for extracting bottom-up information.

Another limitation of the proposed approach relates to the accuracy of the employed object detector.
More specifically, we observe that despite the fact that the objects derived by our approach focus on the area where the event is taking place and explains well event classifier's decision, their labels are not always correct (e.g. see the red colored bars in the WiD barplots of Figs. \ref{fig:shovellingSnow}, \ref{fig:vidExplanation}).
This limitation in the provided explanations is attributed to the imperfection of the object detector.
Nevertheless, we observe that our WiD-based explanation approach highlights the detected objects that are most-related to the recognized event, and which are usually more accurately labeled by the object detector; in this way, it realizes a sort of an error-correcting mechanism on the object detection results (e.g. compare the left and right barplots in Figs. \ref{fig:shovellingSnow}, \ref{fig:vidExplanation}, depicting the objects detected with the highest WiD and DoC values, respectively).
To address the object-detector accuracy limitation, we plan on experimenting with newer object detectors (e.g. \cite{Li_WFIoT_2020,Wang_ICCV_2021}), aiming to further improve the overall accuracy and efficiency of ViGAT as well as the quality of the produced object-level explanations.

\section{Conclusion}
\label{sec:Conclusions}

We presented a new pure-attention bottom-up method for video event recognition, composed of three GAT blocks to process effectively both bottom-up (i.e. object) and frame-level information.
Moreover, utilizing the learned adjacency matrices at the corresponding GAT blocks, WiD-based explanation criteria at object- and frame-level were proposed.
Experimental results on three large, popular datasets showed that the proposed approach achieves state-of-the-art event recognition performance and at the same time provides powerful explanations for the decisions of the model.

As future work, we plan to investigate techniques towards optimizing further the efficiency of ViGAT, for instance, techniques for discarding early in the processing pipeline the objects/frames less correlated with the depicted event, similarly to \cite{GhodratiCVPR2021}; and investigate the utilization of more efficient object detectors and network backbones, such as \cite{Li_WFIoT_2020,Wang_ICCV_2021,Zhang_NIPS_2021},
as well as alternative frame sampling strategies \cite{LiCVPR21,Meng20}.


\begin{thebibliography}{100}
\providecommand{\url}[1]{#1}
\csname url@samestyle\endcsname
\providecommand{\newblock}{\relax}
\providecommand{\bibinfo}[2]{#2}
\providecommand{\BIBentrySTDinterwordspacing}{\spaceskip=0pt\relax}
\providecommand{\BIBentryALTinterwordstretchfactor}{4}
\providecommand{\BIBentryALTinterwordspacing}{\spaceskip=\fontdimen2\font plus
\BIBentryALTinterwordstretchfactor\fontdimen3\font minus
  \fontdimen4\font\relax}
\providecommand{\BIBforeignlanguage}[2]{{%
\expandafter\ifx\csname l@#1\endcsname\relax
\typeout{** WARNING: IEEEtran.bst: No hyphenation pattern has been}%
\typeout{** loaded for the language `#1'. Using the pattern for}%
\typeout{** the default language instead.}%
\else
\language=\csname l@#1\endcsname
\fi
#2}}
\providecommand{\BIBdecl}{\relax}
\BIBdecl

\bibitem{JiangIEEEAccess2022}
J.~Jiang and Y.~Zhang, ``An improved action recognition network with temporal
  extraction and feature enhancement,'' \emph{{IEEE} Access}, vol.~10, pp.
  13\,926--13\,935, 2022.

\bibitem{VieiraIEEEAccess2022}
J.~C. Vieira, A.~Sartori, S.~F. Stefenon, F.~L. Perez, G.~S. de~Jesus, and
  V.~R.~Q. Leithardt, ``Low-cost {CNN} for automatic violence recognition on
  embedded system,'' \emph{{IEEE} Access}, vol.~10, pp. 25\,190--25\,202, 2022.

\bibitem{MemonIEEEAccess2021}
F.~A. Memon, U.~A. Khan, A.~Shaikh, A.~Alghamdi, P.~Kumar, and M.~Alrizq,
  ``Predicting actions in videos and action-based segmentation using deep
  learning,'' \emph{{IEEE} Access}, vol.~9, pp. 106\,918--106\,932, 2021.

\bibitem{WangCASVT2021}
T.~Wang, H.~Zheng, M.~Yu, Q.~Tian, and H.~Hu, ``Event-centric hierarchical
  representation for dense video captioning,'' \emph{{IEEE} Trans. Circuits
  Syst. Video Technol.}, vol.~31, no.~5, pp. 1890--1900, May 2021.

\bibitem{GkalelisCVPR2021}
N.~Gkalelis, A.~Goulas, D.~Galanopoulos, and V.~Mezaris, ``{O}bject{G}raphs:
  Using objects and a graph convolutional network for the bottom-up recognition
  and explanation of events in video,'' in \emph{Proc. IEEE/CVF CVPRW}, Jun.
  2021, pp. 3375--3383.

\bibitem{GhodratiCVPR2021}
A.~Ghodrati, B.~E. Bejnordi, and A.~Habibian, ``{FrameExit}: Conditional early
  exiting for efficient video recognition,'' in \emph{Proc. IEEE/CVF CVPR},
  Virtual Event, Jun. 2021, pp. 15\,608--15\,618.

\bibitem{Meng20}
Y.~Meng, C.~Lin, R.~Panda, P.~Sattigeri, L.~Karlinsky \emph{et~al.},
  ``{AR-Net}: Adaptive frame resolution for efficient action recognition,'' in
  \emph{Proc. ECCV}, Glasgow, UK, Aug. 2020, pp. 86--104.

\bibitem{QiTCSVT2020}
M.~Qi, Y.~Wang, J.~Qin, A.~Li, J.~Luo, and L.~Van~Gool, ``{stagNet}: An
  attentive semantic {RNN} for group activity and individual action
  recognition,'' \emph{{IEEE} Trans. Circuits Syst. Video Technol.}, vol.~30,
  no.~2, pp. 549--565, Feb. 2020.

\bibitem{WangECCV18}
X.~Wang and A.~Gupta, ``Videos as space-time region graphs,'' in \emph{Proc.
  ECCV}, vol. 11209, Munich, Germany, Sep. 2018, pp. 413--431.

\bibitem{Zhang_2021_ICCV}
Y.~Zhang, X.~Li, C.~Liu, B.~Shuai, Y.~Zhu, B.~Brattoli, H.~Chen, I.~Marsic, and
  J.~Tighe, ``{VidTr}: Video transformer without convolutions,'' in \emph{Proc.
  IEEE/CVF ICCV}, Oct. 2021, pp. 13\,577--13\,587.

\bibitem{bertasius_2021_ICML}
G.~Bertasius, H.~Wang, and L.~Torresani, ``Is space-time attention all you need
  for video understanding?'' in \emph{Proc. ICML}, Addis Ababa, Ethiopia, Jul.
  2021.

\bibitem{Arnab_2021_ICCV}
A.~Arnab, M.~Dehghani, G.~Heigold, C.~Sun, M.~Lu\v{c}i\'c, and C.~Schmid,
  ``{ViViT}: A video vision transformer,'' in \emph{Proc. IEEE/CVF ICCV}, Oct.
  2021, pp. 6836--6846.

\bibitem{fan2021multiscale}
H.~Fan, B.~Xiong, K.~Mangalam, Y.~Li, Z.~Yan \emph{et~al.}, ``Multiscale vision
  transformers,'' in \emph{Proc. IEEE/CVF ICCV}, Virtual Event, Oct. 2021, pp.
  6804--6815.

\bibitem{Girdhar_2021_ICCV}
R.~Girdhar and K.~Grauman, ``Anticipative video transformer,'' in \emph{Proc.
  IEEE/CVF ICCV}, Oct. 2021, pp. 13\,505--13\,515.

\bibitem{Sydorov_2019_BMVC}
V.~Sydorov, K.~Alahari, and C.~Schmid, ``Focused attention for action
  recognition,'' in \emph{Proc. BMVC}, Cardiff, UK, Sep. 2019.

\bibitem{Chi_CVPR_2020}
L.~Chi, Z.~Yuan, Y.~Mu, and C.~Wang, ``Non-local neural networks with grouped
  bilinear attentional transforms,'' in \emph{Proc. IEEE/CVF CVPR}, Seattle,
  WA, USA, Jun. 2020, pp. 11\,801--11\,810.

\bibitem{LiCVPR21}
H.~Li, Z.~Wu, A.~Shrivastava, and L.~S. Davis, ``{2D} or not {2D}? adaptive
  {3D} convolution selection for efficient video recognition,'' in \emph{Proc.
  IEEE/CVF CVPR}, Virtual Event, Jun. 2021, pp. 6155--6164.

\bibitem{Li_2022_WACV}
X.~Li, C.~Liu, B.~Shuai, Y.~Zhu, H.~Chen, and J.~Tighe, ``{NUTA:} non-uniform
  temporal aggregation for action recognition,'' in \emph{Proc. IEEE/CVF WACV},
  Waikoloa, HI, USA, Jan. 2022, pp. 827--836.

\bibitem{WangCVPR22}
Y.~Wang, Y.~Yue, Y.~Lin, H.~Jiang, Z.~Lai, V.~Kulikov, N.~Orlov, H.~Shi, and
  G.~Huang, ``{AdaFocus V2}: End-to-end training of spatial dynamic networks
  for video recognition,'' in \emph{Proc. IEEE/CVF CVPR}, New Orleans,
  Louisiana, USA, Jun. 2022.

\bibitem{TsotsosArtifIntel1995}
J.~K. Tsotsos, S.~M. Culhane, W.~Y. {Kei Wai}, Y.~Lai, N.~Davis, and F.~Nuflo,
  ``Modeling visual attention via selective tuning,'' \emph{Artificial
  Intelligence}, vol.~78, no.~1, pp. 507--545, Oct. 1995.

\bibitem{IttiPAMI1998}
L.~Itti, C.~Koch, and E.~Niebur, ``A model of saliency-based visual attention
  for rapid scene analysis,'' \emph{{IEEE} Trans. Pattern Anal. Mach. Intell.},
  vol.~20, no.~11, pp. 1254--1259, 1998.

\bibitem{NelissenScience2005}
K.~Nelissen, G.~Luppino, W.~Vanduffel, G.~Rizzolatti, and G.~A. Orban,
  ``Observing others: Multiple action representation in the frontal lobe,''
  \emph{Science}, vol. 310, no. 5746, pp. 332--336, 2005.

\bibitem{GuptaPAMI2009}
A.~Gupta, A.~Kembhavi, and L.~S. Davis, ``Observing human-object interactions:
  Using spatial and functional compatibility for recognition,'' \emph{{IEEE}
  Trans. Pattern Anal. Mach. Intell.}, vol.~31, no.~10, pp. 1775--1789, Oct.
  2009.

\bibitem{ZacksJEPG2001}
J.~M. Zacks, B.~Tversky, and G.~Iyer, ``Perceiving, remembering, and
  communicating structure in events,'' \emph{J. Exp. Psychol. Gen.}, vol. 130,
  no.~1, pp. 29--58, Mar. 2001.

\bibitem{KurbyTCS2008}
C.~A. Kurby and M.~Z. Jeffrey, ``Segmentation in the perception and memory of
  events,'' \emph{Trends Cogn. Sci.}, vol.~12, no.~2, pp. 72--79, Mar. 2008.

\bibitem{JiCVPR2020}
J.~Ji, R.~Krishna, L.~Fei-Fei, and J.~C. Niebles, ``Action {G}enome: Actions as
  compositions of spatio-temporal scene graphs,'' in \emph{Proc. IEEE/CVF
  CVPR}, Jun. 2020, pp. 10\,236--10\,247.

\bibitem{He_2017_ICCV}
K.~He, G.~Gkioxari, P.~Dollar, and R.~Girshick, ``Mask {R-CNN},'' in
  \emph{Proc. ICCV}, Venice, Italy, Oct. 2017, pp. 2980--2988.

\bibitem{Vaswani_2017_NIPS}
A.~Vaswani, N.~Shazeer, N.~Parmar, J.~Uszkoreit, L.~Jones, A.~N. Gomez,
  L.~Kaiser, and I.~Polosukhin, ``Attention is all you need,'' in \emph{Proc.
  NIPS}, Long Beach, California, USA, 2017, pp. 6000--6010.

\bibitem{WangCVPR2018}
X.~Wang, R.~Girshick, A.~Gupta, and K.~He, ``Non-local neural networks,'' in
  \emph{Proc. IEEE/CVF CVPR}, Salt Lake City, Utah, USA, Jun. 2018, pp.
  7794--7803.

\bibitem{Kaiyu_NIPS_2018}
K.~Yue, M.~Sun, Y.~Yuan, F.~Zhou, E.~Ding, and F.~Xu, ``Compact generalized
  non-local network,'' in \emph{Proc. NIPS}, Montr{\'{e}}al, Canada, Dec. 2018,
  pp. 6511--6520.

\bibitem{HeCVPR2016}
K.~He, X.~Zhang, S.~Ren, and J.~Sun, ``Deep residual learning for image
  recognition,'' in \emph{Proc. IEEE/CVF CVPR}, Las Vegas, NV, USA, Jun. 2016,
  pp. 770--778.

\bibitem{YangCVPR2020}
J.~{Yang}, W.~S. {Zheng}, Q.~{Yang}, Y.~C. {Chen}, and Q.~{Tian},
  ``Spatial-temporal graph convolutional network for video-based person
  re-identification,'' in \emph{Proc. IEEE/CVF CVPR}, Seattle, WA, USA, Jun.
  2020, pp. 3286--3296.

\bibitem{HochreiterNeuralComp97}
S.~Hochreiter and J.~Schmidhuber, ``Long short-term memory,'' \emph{Neural
  Comput.}, vol.~9, no.~8, pp. 1735--1780, 1997.

\bibitem{InanICLR2017}
H.~Inan, K.~Khosravi, and R.~Socher, ``Tying word vectors and word classifiers:
  {A} loss framework for language modeling,'' in \emph{Proc. ICLR}, Toulon,
  France, Apr. 2017.

\bibitem{Lan_2020_ICLR}
Z.~Lan, M.~Chen, S.~Goodman, K.~Gimpel, P.~Sharma, and R.~Soricut, ``{ALBERT:}
  {A} lite {BERT} for self-supervised learning of language representations,''
  in \emph{Proc. ICLR}, Addis Ababa, Ethiopia, Apr. 2020.

\bibitem{GuohaoICML2021}
G.~Li, M.~M{\"u}ller, B.~Ghanem, and V.~Koltun, ``Training graph neural
  networks with 1000 layers,'' in \emph{Proc. ICML}, vol. 139, Jul. 2021, pp.
  6437--6449.

\bibitem{FCVID}
Y.-G. Jiang, Z.~Wu, J.~Wang, X.~Xue, and S.-F. Chang, ``Exploiting feature and
  class relationships in video categorization with regularized deep neural
  networks,'' \emph{{IEEE} Trans. Pattern Anal. Mach. Intell.}, vol.~40, no.~2,
  pp. 352--364, 2018.

\bibitem{XieECCV18}
S.~Xie, C.~Sun, J.~Huang, Z.~Tu, and K.~Murphy, ``Rethinking spatiotemporal
  feature learning: Speed-accuracy trade-offs in video classification,'' in
  \emph{Proc. ECCV}, vol. 11219, Munich, Germany, Sep. 2018, pp. 318--335.

\bibitem{caba2015activitynet}
F.~C. Heilbron, V.~Escorcia, B.~Ghanem, and J.~C. Niebles, ``Activity{N}et: A
  large-scale video benchmark for human activity understanding,'' in
  \emph{Proc. IEEE/CVF CVPR}, 2015, pp. 961--970.

\bibitem{ZhuArxiv2020}
Y.~Zhu, X.~Li, C.~Liu, M.~Zolfaghari, Y.~Xiong, C.~Wu, Z.~Zhang, J.~Tighe,
  R.~Manmatha, and M.~Li, ``A comprehensive study of deep video action
  recognition,'' \emph{CoRR}, vol. abs/2012.06567, 2020.

\bibitem{PareekASO2021}
P.~Pareek and A.~Thakkar, ``A survey on video-based human action recognition:
  recent updates, datasets, challenges, and applications,'' \emph{Artificial
  Intelligence Review}, vol.~54, pp. 2259--2322, 2021.

\bibitem{Simonyan14}
K.~Simonyan and A.~Zisserman, ``Two-stream convolutional networks for action
  recognition in videos,'' in \emph{Proc. NIPS}, vol.~1, Montreal, Canada,
  2014, p. 568–576.

\bibitem{WangECCV16}
L.~Wang, Y.~Xiong, Z.~Wang, Y.~Qiao, D.~Lin, X.~Tang, and L.~V. Gool,
  ``Temporal segment networks: Towards good practices for deep action
  recognition,'' in \emph{Proc. ECCV}, Amsterdam, The Netherlands, Oct. 2016,
  pp. 20--36.

\bibitem{LiAAAI19}
Y.~Li, S.~Song, Y.~Li, and J.~Liu, ``Temporal bilinear networks for video
  action recognition,'' in \emph{Proc. AAAI}, vol.~33, Honolulu, Hawaii, USA,
  Jul. 2019, pp. 8674--8681.

\bibitem{GaoNIPS21}
Z.~Gao, Q.~Wang, B.~Zhang, Q.~Hu, and P.~Li, ``Temporal-attentive covariance
  pooling networks for video recognition,'' in \emph{Proc. NIPS}, vol.~33,
  Virtual Event, Dec. 2021, pp. 13\,587--13\,598.

\bibitem{ZhaoCASVT2018}
S.~Zhao, Y.~Liu, Y.~Han, R.~Hong, Q.~Hu, and Q.~Tian, ``Pooling the
  convolutional layers in deep convnets for video action recognition,''
  \emph{{IEEE} Trans. Circuits Syst. Video Technol.}, vol.~28, no.~8, pp.
  1839--1849, 2018.

\bibitem{Jegou_CVPR_2010}
H.~Jégou, M.~Douze, C.~Schmid, and P.~Pérez, ``Aggregating local descriptors
  into a compact image representation,'' in \emph{Proc. IEEE CVPR}, vol.~34,
  San Francisco, CA, USA, Jun. 2010, pp. 3304--3311.

\bibitem{GirdharCVPR17}
R.~Girdhar, D.~Ramanan, A.~Gupta, J.~Sivic, and B.~C. Russell, ``{ActionVLAD}:
  Learning spatio-temporal aggregation for action classification,'' in
  \emph{Proc. IEEE/CVF CVPR}, Honolulu, HI, USA, Jul. 2017, pp. 3165--3174.

\bibitem{ArandjelovicGTP18}
R.~Arandjelovic, P.~Gron{\'{a}}t, A.~Torii, T.~Pajdla, and J.~Sivic,
  ``Net{VLAD}: {CNN} architecture for weakly supervised place recognition,''
  \emph{{IEEE} Trans. Pattern Anal. Mach. Intell.}, vol.~40, no.~6, pp.
  1437--1451, 2001.

\bibitem{SoltanianAG20}
M.~Soltanian, S.~Amini, and S.~Ghaemmaghami, ``Spatio-temporal {VLAD} encoding
  of visual events using temporal ordering of the mid-level deep semantics,''
  \emph{{IEEE} Trans. Multimedia}, vol.~22, pp. 1769--1784, 2020.

\bibitem{SongTCSVT2020}
X.~Song, C.~Lan, W.~Zeng, J.~Xing, X.~Sun, and J.~Yang, ``Temporal–spatial
  mapping for action recognition,'' \emph{{IEEE} Trans. Circuits Syst. Video
  Technol.}, vol.~30, no.~3, pp. 748--759, Mar. 2020.

\bibitem{XuICMEW18}
Z.~Xu, L.~Su, S.~Wang, Q.~Huang, and Y.~Zhang, ``{S2L}: Single-streamline for
  complex video event detection,'' in \emph{Proc. IEEE ICMEW}, San Diego, CA,
  USA, 2018.

\bibitem{SunghunECCV18}
S.~Kang, J.~Kim, H.~Choi, S.~Kim, and C.~D. Yoo, ``Pivot correlational neural
  network for multimodal video categorization,'' in \emph{Proc. ECCV}, Munich,
  Germany, 2018, pp. 413--431.

\bibitem{WuCVPR19}
Z.~Wu, C.~Xiong, C.~Ma, R.~Socher, and L.~S. Davis, ``{AdaFrame}: Adaptive
  frame selection for fast video recognition,'' in \emph{Proc. IEEE/CVF CVPR},
  Long Beach, CA, USA, Jun. 2020, pp. 1278--1287.

\bibitem{WuICCV19}
W.~Wu, D.~He, X.~Tan, S.~Chen, and S.~Wen, ``Multi-agent reinforcement learning
  based frame sampling for effective untrimmed video recognition,'' in
  \emph{Proc. IEEE/CVF ICCV}, Jul. 2019, pp. 6222--6231.

\bibitem{GaoCVPR2020}
R.~Gao, T.~Oh, K.~Grauman, and L.~Torresani, ``Listen to look: Action
  recognition by previewing audio,'' in \emph{Proc. IEEE/CVF CVPR}, Seattle,
  WA, USA, Jun. 2020, pp. 10\,454--10\,464.

\bibitem{WuNIPS19}
Z.~Wu, C.~Xiong, Y.~Jiang, and L.~S. Davis, ``{LiteEval}: A coarse-to-fine
  framework for resource efficient video recognition,'' in \emph{Proc. NIPS},
  Vancouver, Canada, 2019, pp. 7778--7787.

\bibitem{Wang_2021_ICCV}
Y.~Wang, Z.~Chen, H.~Jiang, S.~Song, Y.~Han, and G.~Huang, ``Adaptive focus for
  efficient video recognition,'' in \emph{Proc. IEEE/CVF ICCV}, Oct. 2021.

\bibitem{GowdaAAAI21}
S.~N. Gowda, M.~Rohrbach, and L.~Sevilla-Lara, ``Smart frame selection for
  action recognition,'' in \emph{Proc. AAAI}, vol. 35(2), May 2021, pp.
  1451--1459.

\bibitem{TranICCV15}
D.~Tran, L.~Bourdev, R.~Fergus, L.~Torresani, and M.~Paluri, ``Learning
  spatiotemporal features with {3D} convolutional networks,'' in \emph{Proc.
  IEEE/CVF ICCV}, Santiago, Chile, 2015, pp. 4489--4497.

\bibitem{CarreiraCVPR17}
J.~Carreira and A.~Zisserman, ``Quo vadis, action recognition? {A} new model
  and the kinetics dataset,'' in \emph{Proc. IEEE/CVF CVPR}, Honolulu, HI, USA,
  Jul. 2017, pp. 4724--4733.

\bibitem{CrastoCVPR19}
N.~Crasto, P.~Weinzaepfel, K.~Alahari, and C.~Schmid, ``{MARS}:
  Motion-augmented {RGB} stream for action recognition,'' in \emph{Proc.
  IEEE/CVF CVPR}, Long Beach, CA, USA, Jun. 2019, pp. 7874--7883.

\bibitem{HaraCVPR18}
K.~Hara, H.~Kataoka, and Y.~Satoh, ``Can spatiotemporal {3D} {CNN}s retrace the
  history of {2D} {CNN}s and {I}mage{N}et?'' in \emph{Proc. IEEE/CVF CVPR},
  Salt Lake City, Utah, USA, Jun. 2018, pp. 6546--6555.

\bibitem{KimCVPR20}
J.~Kim, S.~Cha, D.~Wee, S.~Bae, and J.~Kim, ``Regularization on
  spatio-temporally smoothed feature for action recognition,'' in \emph{Proc.
  IEEE/CVF CVPR}, Seattle, WA, USA, Jun. 2020, pp. 12\,100--12\,109.

\bibitem{FeichtenhoferICCV2020}
C.~Feichtenhofer, H.~Fan, J.~Malik, and K.~He, ``{SlowFast} networks for video
  recognition,'' in \emph{Proc. IEEE/CVF ICCV}, Seoul, Korea (South), Oct.
  2019, pp. 6201--6210.

\bibitem{HaIEEEAccess2021}
M.-H. Ha and O.~T.-C. Chen, ``Deep neural networks using residual fast-slow
  refined highway and global atomic spatial attention for action recognition
  and detection,'' \emph{{IEEE} Access}, vol.~9, pp. 164\,887--164\,902, 2021.

\bibitem{LeeIEEEAccess2021}
Y.~Lee, H.-I. Kim, K.~Yun, and J.~Moon, ``Diverse temporal aggregation and
  depthwise spatiotemporal factorization for efficient video classification,''
  \emph{{IEEE} Access}, vol.~9, pp. 163\,054--163\,064, 2021.

\bibitem{ZhouTMM2022}
J.~Zhou, Z.~Fu, Q.~Huang, Q.~Liu, and Y.~Wang, ``{LgNet}: A local-global
  network for action recognition and beyond,'' \emph{{IEEE} Trans. Multimedia},
  pp. 1--14, Feb. 2022.

\bibitem{ZengIEEEAccess2021}
Q.~Zeng, M.~O. Tezcan, and J.~Konrad, ``Dynamic equilibrium module for action
  recognition,'' \emph{{IEEE} Access}, vol.~9, pp. 168\,015--168\,025, 2021.

\bibitem{VuIEEEAccess2021}
D.-Q. Vu, N.~Le, and J.-C. Wang, ``Teaching yourself: A self-knowledge
  distillation approach to action recognition,'' \emph{{IEEE} Access}, vol.~9,
  pp. 105\,711--105\,723, 2021.

\bibitem{TranCVPR2018}
D.~Tran, H.~Wang, L.~Torresani, J.~Ray, Y.~LeCun, and M.~Paluri, ``A closer
  look at spatiotemporal convolutions for action recognition,'' in \emph{Proc.
  IEEE/CVF CVPR}, Salt Lake City, UT, USA, Jun. 2018, pp. 6450--6459.

\bibitem{FayyazCVPR21}
M.~Fayyaz, E.~B. Rad, A.~Diba, M.~Noroozi, E.~Adeli, L.~V. Gool, and J.~Gall,
  ``{3D CNNs} with adaptive temporal feature resolutions,'' in \emph{Proc.
  IEEE/CVF CVPR}, Virtual Event, Jun. 2021, pp. 4731--4740.

\bibitem{KorbarICCV9}
B.~Korbar, D.~Tran, and L.~Torresani, ``{SCSampler}: Sampling salient clips
  from video for efficient action recognition,'' in \emph{Proc. IEEE/CVF ICCV},
  Seoul, Korea (South), Oct.\slash Nov. 2019, pp. 6231--6241.

\bibitem{YuanCVPR2020}
C.~Wu, R.~B. Girshick, K.~He, C.~Feichtenhofer, and P.~Kr{\"{a}}henb{\"{u}}hl,
  ``A multigrid method for efficiently training video models,'' in \emph{Proc.
  IEEE/CVF CVPR}, Seattle, WA, USA, Jun. 2020, pp. 150--159.

\bibitem{FeichtenhoferCVPR2020}
C.~Feichtenhofer, ``{X3D:} expanding architectures for efficient video
  recognition,'' in \emph{Proc. IEEE/CVF CVPR}, Seattle, WA, USA, Jun. 2020,
  pp. 200--210.

\bibitem{Tan_ICML_2019}
M.~Tan and Q.~Le, ``{E}fficient{N}et: Rethinking model scaling for
  convolutional neural networks,'' in \emph{Proc. ICML}, vol.~97, Long Beach,
  CA, USA, Jun. 2019, pp. 6105--6114.

\bibitem{YangICCV21}
G.~Yang, H.~Tang, M.~Ding, N.~Sebe, and E.~Ricci, ``Transformer-based attention
  networks for continuous pixel-wise prediction,'' in \emph{Proc. IEEE/CVF
  ICCV}, Montreal, QC, Canada, Oct. 2021, pp. 16\,249--16\,259.

\bibitem{DosovitskiyICLR21}
A.~Dosovitskiy, L.~Beyer, A.~Kolesnikov, D.~Weissenborn, X.~Zhai \emph{et~al.},
  ``An image is worth 16x16 words: Transformers for image recognition at
  scale,'' in \emph{Proc. ICLR}, Virtual Event, May 2021.

\bibitem{LuoCASVT2021}
H.~Luo, G.~Lin, Y.~Yao, Z.~Tang, Q.~Wu, and X.~Hua, ``Dense semantics-assisted
  networks for video action recognition,'' \emph{{IEEE} Trans. Circuits Syst.
  Video Technol.}, 2021.

\bibitem{WuCVPR2019}
C.~Wu, C.~Feichtenhofer, H.~Fan, K.~He, P.~Kr{\"{a}}henb{\"{u}}hl, and R.~B.
  Girshick, ``Long-term feature banks for detailed video understanding,'' in
  \emph{Proc. IEEE/CVF CVPR}, Long Beach, CA, USA, Jun. 2019, pp. 284--293.

\bibitem{NigamCASVT2021}
N.~Nigam, T.~Dutta, and H.~P. Gupta, ``{FactorNet}: Holistic actor, object and
  scene factorization for action recognition in videos,'' \emph{{IEEE} Trans.
  Circuits Syst. Video Technol.}, 2021.

\bibitem{ZhangCVPR2019}
J.~Zhang, K.~J. Shih, A.~Elgammal, A.~Tao, and B.~Catanzaro, ``Graphical
  contrastive losses for scene graph parsing,'' in \emph{Proc. IEEE/CVF CVPR},
  Long Beach, CA, USA, Jun. 2019, pp. 11\,535--11\,543.

\bibitem{DaiNips2016}
J.~Dai, Y.~Li, K.~He, and J.~Sun, ``R-{FCN}: Object detection via region-based
  fully convolutional networks,'' in \emph{Proc. NIPS}, Barcelona, Spain, Jun.
  2016, pp. 379--387.

\bibitem{MunkresJSIAM1957}
J.~Munkres, ``Low-cost {CNN} for automatic violence recognition on embedded
  system,'' \emph{Algorithms for the Assignment and Transportation Problems},
  vol.~5, no.~1, pp. 32--38, 1957.

\bibitem{SudhakarSPJJK21}
M.~Sudhakar, S.~Sattarzadeh, K.~N. Plataniotis, J.~Jang, Y.~Jeong, and H.~Kim,
  ``{Ada-Sise}: Adaptive semantic input sampling for efficient explanation of
  convolutional neural networks,'' in \emph{Proc. IEEE ICASSP}, Toronto, ON,
  Canada, Jun. 2021, pp. 1715--1719.

\bibitem{YuanExplainability}
H.~Yuan, H.~Yu, S.~Gui, and S.~Ji, ``Explainability in graph neural networks: A
  taxonomic survey,'' \emph{CoRR}, vol. 2012.15445, 2020.

\bibitem{YingNIPS2019}
R.~Ying, D.~Bourgeois, J.~You, M.~Zitnik, and J.~Leskovec, ``{GNNExplainer}:
  Generating explanations for graph neural networks,'' in \emph{Proc. NIPS},
  Vancouver, Canada, 2019, pp. 9240--9251.

\bibitem{AgarwalPMLR22}
C.~Agarwal, M.~Zitnik, and H.~Lakkaraju, ``Probing {GNN} explainers: {A}
  rigorous theoretical and empirical analysis of {GNN} explanation methods,''
  in \emph{Proc. IEEE/CVF MLR}, vol. 151, Virtual Event, Mar. 2022, pp.
  8969--8996.

\bibitem{VuNIPS20}
M.~N. Vu and M.~T. Thai, ``{PGM-Explainer}: Probabilistic graphical model
  explanations for graph neural networks,'' in \emph{Proc. NIPS}, Vancouver,
  Canada, Dec. 2020.

\bibitem{FunkeArxiv2021}
T.~Funke, M.~Khosla, and A.~Anand, ``Zorro: Valid, sparse, and stable
  explanations in graph neural networks,'' \emph{CoRR}, vol. 2105.08621, 2021.

\bibitem{PopeCVPR19}
P.~E. Pope, S.~Kolouri, M.~Rostami, C.~E. Martin, and H.~Hoffmann,
  ``Explainability methods for graph convolutional neural networks,'' in
  \emph{Proc. IEEE/CVF CVPR}, Vancouver, Canada, Dec. 2019, pp.
  10\,764--10\,773.

\bibitem{SaurabhWACV20}
S.~Desai and H.~G. Ramaswamy, ``Ablation-{CAM}: Visual explanations for deep
  convolutional network via gradient-free localization,'' in \emph{Proc.
  IEEE/CVF WACV}, Mar. 2020, pp. 972--980.

\bibitem{ScarselliIEEETNN2009}
F.~Scarselli, M.~Gori, A.~C. Tsoi, M.~Hagenbuchner, and G.~Monfardini, ``The
  graph neural network model,'' \emph{{IEEE} Trans. Neural Netw.}, vol.~20,
  no.~1, pp. 61--80, Jan. 2009.

\bibitem{Velickovic_ICLR_2018}
P.~Velickovic, G.~Cucurull, A.~Casanova, A.~Romero, P.~Li{\`{o}}, and
  Y.~Bengio, ``Graph attention networks,'' in \emph{Proc. ICLR}, Vancouver, BC,
  Canada, Apr.\slash May 2018.

\bibitem{kipfIclr2017}
T.~N. Kipf and M.~Welling, ``Semi-supervised classification with graph
  convolutional networks,'' in \emph{Proc. ICLR}, Toulon, France, Apr. 2017.

\bibitem{ba2016layer}
J.~L. Ba, J.~R. Kiros, and G.~E. Hinton, ``Layer normalization,'' \emph{CoRR},
  vol. abs/1607.06450, 2016.

\bibitem{LeeIcml2019}
J.~Lee, I.~Lee, and J.~Kang, ``Self-attention graph pooling,'' in \emph{Proc.
  ICML}, Jun. 2015, pp. 3734--3743.

\bibitem{Justin_ICML_2017}
J.~Gilmer, S.~S. Schoenholz, P.~F. Riley, O.~Vinyals, and G.~E. Dahl, ``Neural
  message passing for quantum chemistry,'' in \emph{Proc. ICML}, vol.~70,
  Sydney, NSW, Australia, Aug. 2017, pp. 1263--1272.

\bibitem{KayArxiv2017}
W.~Kay, J.~Carreira, K.~Simonyan, B.~Zhang, C.~Hillier \emph{et~al.}, ``The
  kinetics human action video dataset,'' \emph{CoRR}, vol. abs/1705.06950,
  2017.

\bibitem{renNips2015faster}
S.~Ren, K.~He, R.~Girshick, and J.~Sun, ``Faster {R-CNN}: Towards real-time
  object detection with region proposal networks,'' in \emph{Proc. NIPS},
  vol.~28, 2015.

\bibitem{ILSVRC15}
O.~Russakovsky, J.~Deng, H.~Su, J.~Krause, and S.~Satheesh, ``{ImageNet} large
  scale visual recognition challenge,'' \emph{Int. J. Comput. Vision}, vol.
  115, no.~3, pp. 211--252, 2015.

\bibitem{krishna2017}
R.~Krishna, Y.~Zhu, O.~Groth, J.~Johnson, K.~Hata \emph{et~al.}, ``Visual
  {Genome}: Connecting language and vision using crowdsourced dense image
  annotations,'' \emph{Int. J. Comput. Vision}, vol. 123, no.~1, pp. 32--73,
  May 2017.

\bibitem{ZhangEDS2009}
E.~Zhang and Y.~Zhang, \emph{Average Precision}.\hskip 1em plus 0.5em minus
  0.4em\relax Boston, MA: Springer US, 2009, pp. 192--193.

\bibitem{ChattopadhayWACV2018}
A.~Chattopadhay, A.~Sarkar, P.~Howlader, and V.~N. Balasubramanian,
  ``Grad-{CAM}++: Generalized gradient-based visual explanations for deep
  convolutional networks,'' in \emph{Proc. IEEE WACV}, Mar. 2018, pp. 839--847.

\bibitem{fvcorFlopCount}
``Flop counter for pytorch models,''
  \url{https://github.com/facebookresearch/fvcore/blob/main/docs/flop_count.md},
  accessed: 2022-04-07.

\bibitem{Carion_2020_ECCV}
N.~Carion, F.~Massa, G.~Synnaeve, N.~Usunier, A.~Kirillov, and S.~Zagoruyko,
  ``End-to-end object detection with transformers,'' in \emph{Proc. ECCV},
  Glasgow, UK, Aug. 2020, pp. 213–--229.

\bibitem{Li_WFIoT_2020}
Y.~Li, A.~Dua, and F.~Ren, ``Light-weight {RetinaNet} for object detection on
  edge devices,'' in \emph{Proc. IEEE WF-IoT}, Virtual Event, Jun. 2020.

\bibitem{Zhang_NIPS_2021}
Q.~Zhang and Y.-B. Yang, ``{ResT}: An efficient transformer for visual
  recognition,'' in \emph{Proc. NIPS}, vol.~34, Dec. 2021, pp.
  15\,475--15\,485.

\bibitem{Chen2020}
M.~Chen, Z.~Wei, Z.~Huang, B.~Ding, and Y.~Li, ``Simple and deep graph
  convolutional networks,'' in \emph{Proc. MLR}, vol. 119, Jul. 2020, pp.
  1725--1735.

\bibitem{GkalelisICIP2014}
N.~{Gkalelis}, V.~{Mezaris}, I.~{Kompatsiaris}, and T.~{Stathaki}, ``Video
  event recounting using mixture subclass discriminant analysis,'' in
  \emph{Proc. IEEE ICIP}, Melbourne, Australia, Sep. 2013, pp. 4372--4376.

\bibitem{Wang_ICCV_2021}
C.-Y. Wang, H.-Y.~M. Liao, I.-H. Yeh, Y.-Y. Chuang, and Y.-L. Lin, ``Exploring
  the power of lightweight {YOLOv4},'' in \emph{Proc. IEEE/CVF ICCVW}, Virtual
  Event, Oct. 2021, pp. 779--788.

\end{thebibliography}


\begin{IEEEbiography}[{\includegraphics[width=1in,height=1.25in,clip,keepaspectratio]{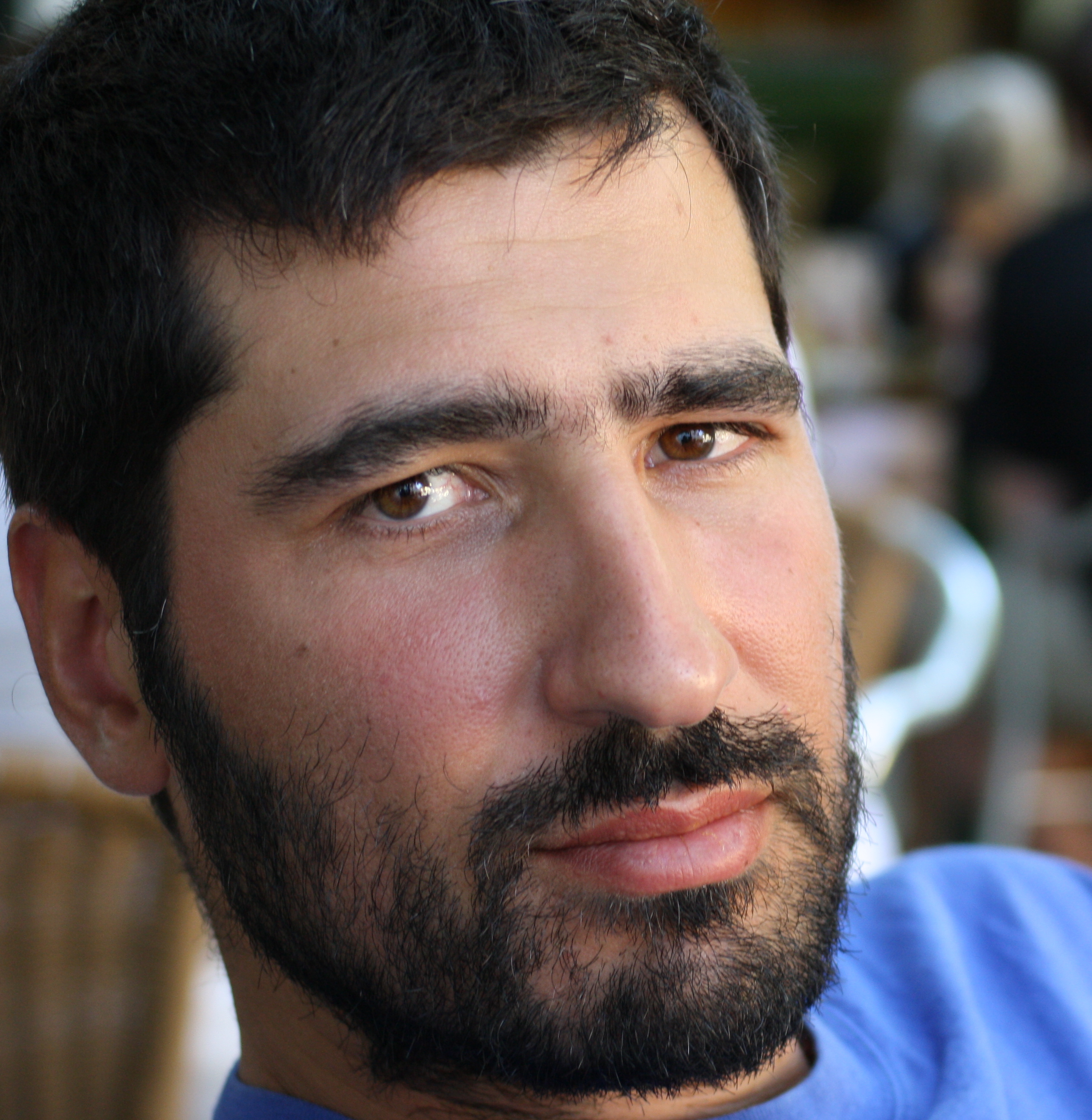}}]
{Nikolaos Gkalelis} received the B.Sc. degree in electronic engineering from the Technological Institute of Thessaloniki, Greece, in 2000, the M.Sc. degree in control systems engineering with distinction from the University of Sheffield, U.K., in 2002, the M.Sc. degree in computational engineering from the University of Erlangen-Nuremberg, Germany, in 2004, and the Ph.D. degree in electrical engineering from Imperial College London, U.K., in 2014. He worked in R\&D for Aristotle University of Thessaloniki, Siemens and ECB. He is currently a Postdoctoral Researcher with the Centre for Research and Technology Hellas, Information Technologies Institute. He has coauthored 4 journal articles, 1 book chapter, 31 conference papers, and 1 patent. His research interests include deep learning, machine learning, video analysis, multimedia understanding and digital signal processing. He serves regularly as a reviewer for many international journals and conferences.
\end{IEEEbiography}

\begin{IEEEbiography}[{\includegraphics[width=1in,height=1.25in,clip,keepaspectratio]{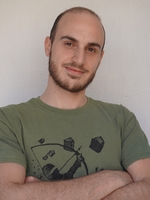}}]
{Dimitrios Daskalakis}
received his B.Sc. degree in electrical and computer engineering from the Aristotle University of Thessaloniki, Greece, in 2021. He is currently a Research Associate with the Center for Research and Technology Hellas, Information Technologies Institute. His research interests include machine learning and deep learning for multimedia understanding as well as video and image analysis.
\end{IEEEbiography}

\begin{IEEEbiography}[{\includegraphics[width=1in,height=1.25in,clip,keepaspectratio]{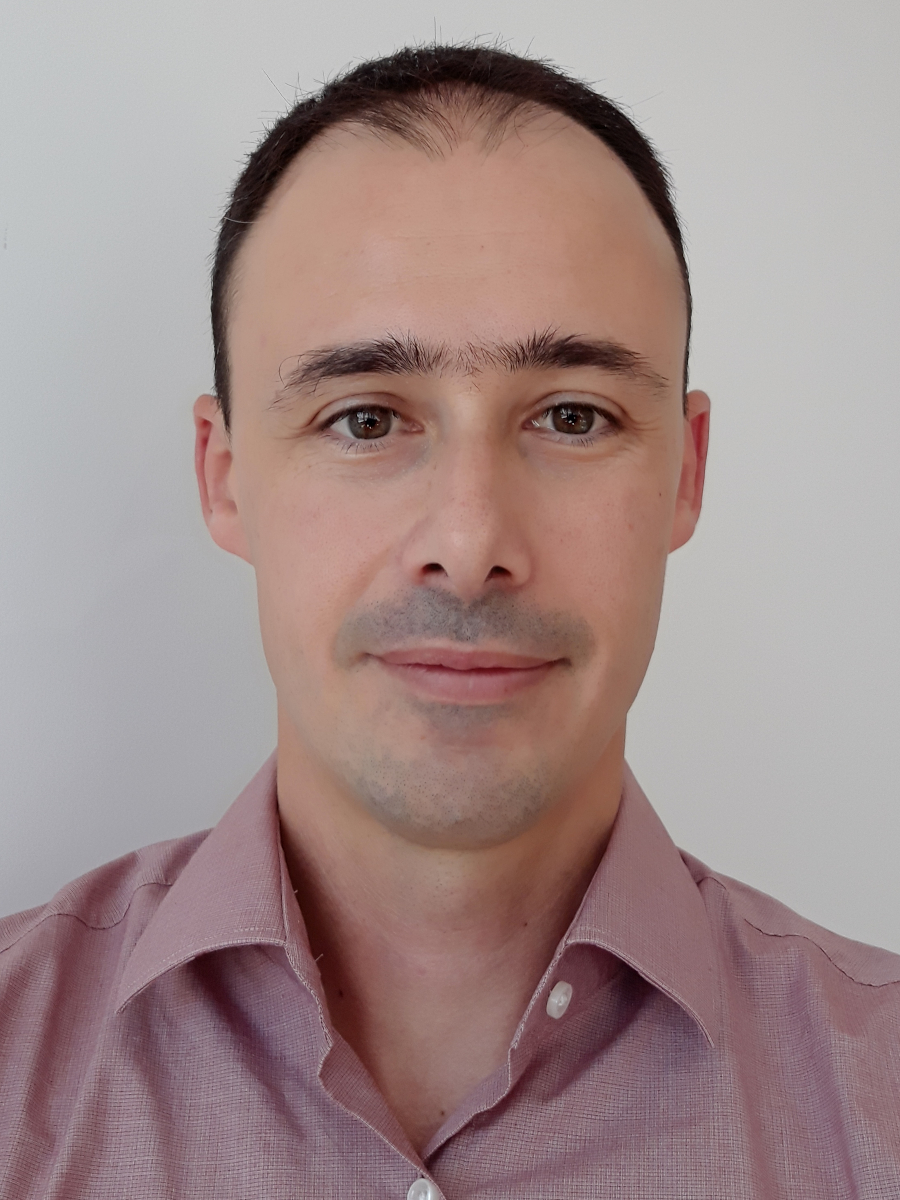}}]
{Vasileios Mezaris} (Senior Member, IEEE) received the B.Sc. and Ph.D. degrees in electrical and computer engineering from the Aristotle University of Thessaloniki, Greece, in 2001 and 2005, respectively. He is currently a Research Director with the Centre for Research and Technology Hellas, Information Technologies Institute. He has coauthored more than 40 journal articles, 20 book chapters, 180 conference papers, and three patents. His research interests include multimedia understanding and artificial intelligence, in particular, image and video analysis and annotation, machine learning and deep learning for multimedia understanding and big data analytics, multimedia indexing and retrieval, and applications of multimedia understanding and artificial intelligence. He serves as a Senior Area Editor for IEEE Signal Processing Letters and served as an Associate Editor for IEEE Transactions on Multimedia.
\end{IEEEbiography}

\EOD

\end{document}